\documentclass[conference]{IEEEtran}
\usepackage{times}

\usepackage[numbers]{natbib}
\usepackage{multicol}
\usepackage[bookmarks=true]{hyperref}
\usepackage{xcolor}
\usepackage{amsmath, amssymb} 
\usepackage{bm}
\usepackage{graphicx}
\usepackage{dsfont}
\usepackage{threeparttable}
\usepackage{amsthm}
\usepackage{subcaption}

\usepackage{algorithm}
\usepackage{algpseudocode}

\usepackage[numbers]{natbib}

\newcommand{\argmax}{\mathop{\rm arg~max}\limits}

\newcommand{\Revise}[1]{{#1}}

\theoremstyle{remark} 

\begin{document}

\title{Rethinking Robustness Assessment: Adversarial Attacks on Learning-based Quadrupedal Locomotion Controllers}

\author{Fan Shi$^{\dagger 123}$, Chong Zhang$^{\dagger 1}$, Takahiro Miki$^{1}$, Joonho Lee$^{1}$, Marco Hutter$^{1}$, Stelian Coros$^{2}$%
\\ {$^{\dagger }$ Equal contribution. Video: \url{https://fanshi14.github.io/me/rss24.html}} 
\\ {$^{1}$ Robotic Systems Lab, ETH Zurich $^{2}$ Computational Robotics Lab, ETH Zurich $^{3}$ ETH AI Center}
}

\maketitle

\begin{abstract}
Legged locomotion has recently achieved remarkable success with the progress of machine learning techniques, especially deep reinforcement learning (RL). Controllers employing neural networks have demonstrated empirical and qualitative robustness against real-world uncertainties, including sensor noise and external perturbations. 
However, formally investigating the vulnerabilities of these locomotion controllers remains a challenge. This difficulty arises from the requirement to pinpoint vulnerabilities across a long-tailed distribution within a high-dimensional, temporally sequential space. 
As a first step towards quantitative verification, we propose a computational method that leverages sequential adversarial attacks to identify weaknesses in learned locomotion controllers.
Our research demonstrates that, even state-of-the-art robust controllers can fail significantly under well-designed, low-magnitude adversarial sequence.
Through experiments in simulation and on the real robot, we validate our approach's effectiveness, and we illustrate how the results it generates can be used to robustify the original policy and offer valuable insights into the safety of these black-box policies.
\end{abstract}

\IEEEpeerreviewmaketitle

\section{Introduction}


Legged robots, especially quadrupedal robots, are expected to free people from dull, dirty, and dangerous ($3$D) tasks, such as transportation, underground inspections, and rescue operations.
However, real-world scenarios with complex terrain, sensor noises, and unexpected perturbations present significant hurdles.
Robust locomotion controllers are essential in this context.

Recent advancements in machine learning, especially in reinforcement learning (RL), have significantly improved the performance of quadrupedal robots.~\cite{lee2020scirob, yang2020multi, miki2022scirob}. Controllers based on neural networks (NN) have demonstrated remarkable robustness, contributing to the success of team CERBERUS in the DARPA Subterranean Challenge~\cite{rslsubt2022}. This competition is known for its demanding mobility tasks in real underground environments~\cite{chung2023subt}. Unlike traditional controllers that rely on explicit dynamics models and online optimization~\cite{jenelten2022tamols, grandia2023perceptive}, neural network controllers reformulate the problem into offline optimization of the control policy through massive trials and
errors in simulation.

\begin{figure}[t]
  \begin{center}
    \includegraphics[clip,  bb= 0 0 600 420,  width=1.0\columnwidth]{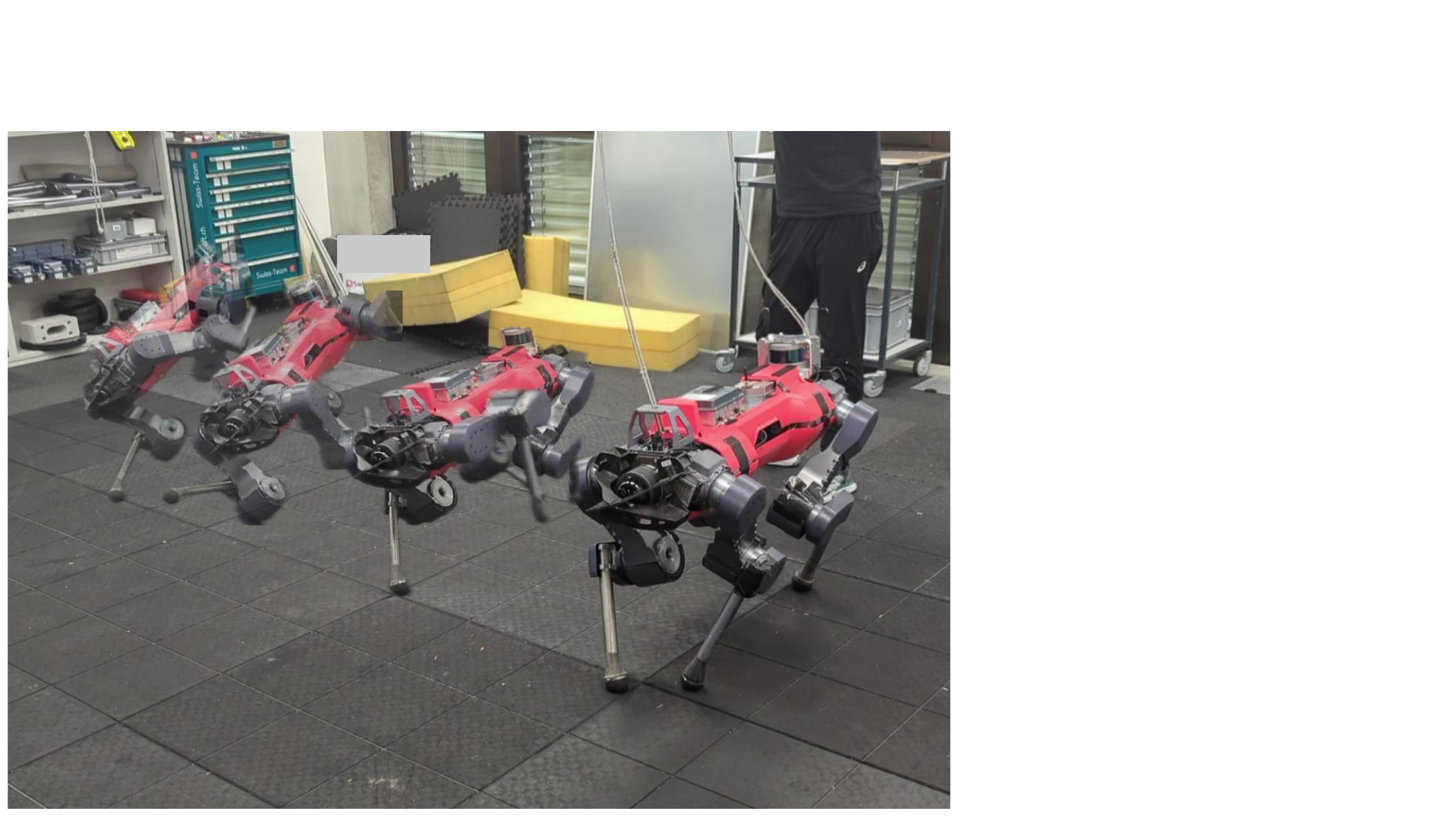}
    \caption{The state-of-the-art robust locomotion policy~\cite{miki2022scirob} can be destabilized even on flat ground when subjected to a sequence of low-magnitude multi-modal adversarial attacks on observations, demonstrating the vulnerabilities in learning-based neural controllers.}
    \label{figure:fall}
  \end{center}
  \vspace{-3mm}
\end{figure}

Despite the rich literature on neural locomotion controllers, there's still a limited understanding of how neural networks operate, making these controllers seem like a "black box" to researchers~\cite{lipton2018mythos}. Recent studies have increasingly focused on the vulnerabilities of neural networks~\cite{croce2021robustbench, abbeel2017adversarial-nn-policy}, highlighting how small, adversarial modifications can significantly impact their performance. These subtle manipulations pose considerable risks in practical applications of neural network systems.

In the context of robot control, adversarial attacks can be formulated as introducing noises or disturbances that compromise the effectiveness of the controllers. For instance, \citet{koren2018ast} proposed to identify specific pedestrian movement patterns that could potentially cause an autonomous vehicle to crash \cite{koren2018ast}. Yet, within the domain of quadrupedal locomotion controllers, validation of such risks with temporally sequential attacks is still missing, highlighting the need of further exploration to understand and mitigate potential vulnerabilities.

Given the high-dimensional and multi-modal nature of the search space for sequential attacks, strategies such as introducing random noise through domain randomization~\cite{tobin2017domain} are insufficient for identifying the subtle vulnerabilities of the controllers. Moreover, policies which are designed to be resilient against such disturbances~\cite{miki2022scirob} present further challenges to discover effective sequential attacks using straightforward methods.
Additionally, adversarial scenarios should be grounded in \textit{realism}, implying they are scenarios that the robot might realistically encounter during its operational tasks. Highly improbable conditions, like disturbances significantly beyond normal environmental challenges, do not contribute effectively to our understanding of a controller's nuanced vulnerabilities. 

Our approach, therefore, is to generate adversaries in a way that is both methodical and informed by field robotics experience. This ensures that the scenarios we explore are not only plausible but also indicative of real-world conditions the robots might face.
Moreover, the identification of the mildest successful adversarial sequence can also indicate the worst-case robustness, providing insights into the system's safety. This comprehensive understanding will facilitate the future large-scale deployment of neural policies, ensuring they are equipped to handle a broad spectrum of operational challenges effectively.
Furthermore, our findings reveal that the robustness of control policies can be notably increased by finetuning them with the adversarial scenarios we've identified, thereby enhancing their ability to withstand unexpected challenges.

Briefly, we identify our contribution as follows:

\begin{enumerate}

\item \textbf{A learning approach} that effectively discovers sequential adversarial attacks on neural quadrupedal locomotion controllers;
\item \textbf{Vulnerability analysis} of state-of-the-art quadrupedal locomotion controllers under multi-modal attacks and demonstrate on an actual robot;
\item \textbf{Closed-loop integration} with adversarial attacks to robustify and analyze the locomotion controllers, complemented by real-world validations.
\end{enumerate}

\section{Related work}

\subsection{Neural Controllers for Legged Robots}

Neural controllers have shown impressive results on legged robots, achieving capabilities such as robust locomotion~\cite{jemin2019scirob, lee2020scirob, yang2020multi, miki2022scirob, carius2020mpc-net, usc2022robust, mit2022rapid, kumar2021rma, li2021reinforcement, DBLP:conf/rss/SiekmannGWFH21, siekmann2021sim}, mimicking animal behaviors~\cite{peng2020quadruped, peng2021amp}, interacting with objects~\cite{shi2021circus, tanjie2020rarl}, and performing dynamic, athletic movements~\cite{hoeller2023parkour, cmu2023parkour}. 
The advantage of using neural policies lies in their ability to handle the control problem's complexity and non-linearity efficiently, offering a significant reduction in computational demand during operation compared to traditional model-based optimization approaches~\cite{jemin2019scirob}. Furthermore, by massive training with varied simulations that include random noises, disturbances, and dynamic changes~\cite{tobin2017domain, xie2021dr}, model-free RL-based controllers can learn to adapt and stabilize the system under a wide range of conditions.

Despite their effectiveness, neural controllers often operate as a "black box" to developers, making it difficult to predict or comprehend their failure scenarios.
This paper investigates how vulnerabilities of NNs affect the worst-case safety of quadrupedal locomotion controllers, and how they can be "patched" to robustify the controllers. 
Compared to another line of recent works using distributional RL to optimize the rewards for the most extreme case with large perturbations during training~\cite{schneider2023learning, li2024learning, shi2023robust}, our work can even identify subtle vulnerabilities of controllers during deployment, requiring no transparency of the controller training.

\subsection{Safety Validation on Locomotion Controller}

To improve the safety of robot controllers, an effective safety validation process is necessary. However, this is nontrivial due to two primary challenges: the limited validation scenarios that may not fully represent the real world’s complexity, and the high computational cost associated with searching for rare failures.

Regarding the scenarios, some of existing works develop standard push-recovery testbeds by applying fixed external perturbations to the legged robot with different control policies \cite{zhangei2023quadruped-tro, weng2023standard-legged-test, weng2023comparability}. \citet{zhang2023generating} generate several terrain templates that is challenging and realistic to test the robustness of locomotion policies. However, these testbeds with fixed settings cover only a very limited range of scenarios. In contrast, real-world situations present more complex challenges.

Regarding the searching efficiency, as the control problems are time-sequential in a high-dimensional space, naive searching methods can be costly due to the ‘curse of dimensionality’~\cite{brundage2020trustworthy-ai-survey}. Hence, optimization-based methods, such as RL, are adopted to speed up the validation process in recent works. For example, in autonomous driving some use RL to learn successful adversarial attacks that uncover the hiding risks \cite{koren2018ast, corso2019augment-adaptive, umich2023nature-av-test}. In the context of quadrupedal robots, a recent work involves using an evolutionary algorithm to identify adversarial attacks targeting joint torques \cite{otomo2022joints-adv-quadruped}. However, scenarios involving extreme joint torque failures are uncommon in real-world settings, and this work's analysis is based on a simplified "Ant" model, lacking validation with actual robots.

\subsection{Robustify Control Policy with Adversarial Samples}
\Revise{Adversarial attacks can be introduced during the training phase to strengthen the resilience of control policies \cite{pattanaik2018robust}. 
For instance, \citet{pan2023grasping-adv} shows that adversarial objects can be used to develop a more robust policy for grasping. 
Similarly, the robust adversarial RL technique \cite{pinto2017rarl} has been effective in enhancing robustness in competitive scenarios, such as active object tracking \cite{zhong2019ad, zhong2021towards, devo2021enhancing}, quadruped object following \cite{tanjie2020rarl}, and simulated humanoid boxing \cite{ucb2019adversarial-two-agents}.}

However, most related studies concentrate on interactions between legged robots and other agents, treating these interactions as self-play \cite{ucb2019adversarial-two-agents, tanjie2020rarl}.
In contrast, our paper models sensor noises, command signals, and perturbation events on the quadrupedal robot as adversarial samples, which are further leveraged to enhance the controller's robustness.
We validate the efficacy of our approach both in simulation and on a actual robot.

\Revise{
In addition to employing well-grounded adversaries based on field experiences, our work designs distinct policy and adversarial loss function, inspired by \cite{zhang2019robustness_accuracy, robey2023nonzerosum, croce2020ensemble_adv, peng2021amp, peng2022ase, marko2023adv-motor-prior, li2023learning, li2023versatile}. We direct the adversary to prioritize safety violations rather than the typical zero-sum formulation that seeks to minimize the policy's training rewards. This focus is crucial for safety-critical systems, such as legged robots, where rewards are often dominated by performance objectives (e.g., velocity tracking accuracy), and minimizing these does not necessarily result in catastrophic failures.
}

\section{Preliminaries}

\subsection{Quadrupedal Neural Locomotion Policy}

Neural controllers of quadrupedal locomotion map their observations to the action space which can be joint targets tracked by PD controllers~\cite{rudin2022corl} or parameters that are convertible to joint-level outputs~\cite{lee2020scirob, miki2022scirob}. The observations typically come from onboard sensors, such as IMU, joint encoders, or height scan from lidars and depth cameras.


\subsection{Reinforcement Learning}
The robot discrete-time control problem could be formulated as a Markov Decision Process (MDP) to maximize the total task reward as Eq. \ref{eq:rl}.
It is defined by $(S, A, R, P, \gamma$), including states, actions, reward function, state transition probabilities, and discounter factor.
The policy $\pi$ is a distribution over action given states as $\pi (a_t \in A |s_t \in S)$ and receive reward $r_t \in R$ through interaction with the environment, following the transition probability $P(s_{t+1} | s_t, a_t)$.
In deep RL, the policy is parameterized by neural network parameters $\theta$, which is optimized towards

\begin{equation}
  \label{eq:rl}
\theta^{*} = \argmax_{\theta} \mathbb{E}_{\pi_{\theta}}\left[
      \sum_{t=0}^{\infty} \gamma^{t} r_{t}
      \right].
\end{equation}

Considering the large number of samples required for learning, many works train control policies with the parallel simulation pipeline~\cite{hwangbo2018per, makoviychuk2021isaac} to speed up learning, avoid hardware damage during sample collection, or obtain privileged information.

\subsection{Domain Randomization}
Robots are expected to function in the real world. 
However, there is a large gap between simulation and the real world, necessitating techniques such as domain randomization (DR)~\cite{tobin2017domain} to bridge the gap. DR randomizes the environment settings, noises, and perturbations in simulation so that the learned policy can generalize to the real world. Despite the efficacy of DR for sim-to-real transfer, we find that the learned neural policy still exhibits vulnerabilities.

\subsection{Definition of Robust Policy}
\label{pre:robust-def}

In this paper, we focus on deliberately inducing failures in quadrupedal controllers, scenarios usually circumvented as termination conditions during the training of locomotion policies. Examples include the robot falling over or experiencing collisions at its base, where weight and sensors are heavily centralized. The robustness of the policy is defined by its capacity to prevent such failures in the face of adversarial attacks.


\section{Learning Adversarial Attacks on Quadrupedal Locomotion}

\subsection{Adversarial Space}

Selecting the appropriate attacks for a robot controller involves determining an adversarial space that effectively reveals the vulnerabilities of a locomotion policy. The initial step is to identify realistic adversarial scenarios that a robot is likely to face during actual deployment. We focus on sensor noise, external disturbances, and hazardous user commands as the main sources of failure. Consequently, we identify the observation space, the command space, and the perturbation space as potentially vulnerable to attacks.

For each adversarial space, it's important to define appropriate ranges and rates of change. Extremely large attacks, such as external forces of $10,000$ N on a $50$-kg robot, or user commands changing at a rate of $100$ m/s$^2$, are unrealistic and would not provide useful insights into a policy's robustness under typical operational conditions. Hence, the selection of these ranges and their rates of change must be informed by practical experience in field robotics. In our experiments, we carefully adjust the boundaries and rates of change of the generated adversarial inputs to ensure they remain within realistic limits.

Regarding the command space, locomotion policies typically track velocity commands from the user or the navigation module, namely the linear velocity in the $x-y$ plane and the angular velocity around the $z$-axis. We attack the command space by giving malicious commands to be tracked.

Regarding the observation space, most of the existing locomotion policies use gathered base rotation and joint state data from the state estimator and joint encoders. Modern joint encoders exhibit negligible errors~\cite{encoder-accuracy}, while the pose estimation~\cite{bloesch2013state, bloesch2017estimator} continues to demonstrate significant errors caused by the noisy IMU sensor and modelling mismatch, as shown in Fig.~\ref{figure:state-estimator-GT}.
Therefore, we ignore joint encoder errors in this paper, and limit the attacks to the rotation errors obtained from the robot's state estimator.

Regarding the perturbation space, we attack the robot with external forces that are not observed by the controller.
These forces can be applied on any body part, exemplified by Sec.~\ref{case:blind} where forces target the base to simulate pushes and payload, and by Sec.~\ref{case:darpa} where forces target the feet to simulate stumbling and slippage. 
Note that we can only control the forces in simulation. Therefore, we do not apply perturbations during the real-world validation of the attack. 


\subsection{RL-based Attack Learning with Lipschitz Regularization}
\label{sec:attack-lip}

To efficiently search for failures that are time-sequential, we train adversarial RL policies that generate attacks based on the robot's state. For an adversary policy $\pi_{\theta_{adv}}$ parameterized by $\theta_{adv}$, we optimize the parameters towards
\begin{equation}
  \label{eq:rl-adv}
\resizebox{1.0\hsize}{!}{%
    $ \theta_{adv}^{*} = \argmax_{\theta_{adv}} \mathop{\mathbb{E}}\limits_{\pi_{\theta_{adv}}}\left[\sum\limits_{t=0}^{\infty}  \gamma^{t} (-\mathds{1}_{alive} + r_t^{aux}) \right] -\lambda\cdot\prod\limits_i\lVert \theta_{adv}^i \rVert_{\infty},$
    }%
\end{equation}
where $\mathds{1}_{alive}$ constantly penalizes the adversarial policy if the robot doesn't fail, $r_t^{aux}$ is the summation of auxiliary rewards that facilitate exploration, $\lambda$ is the coefficient of Lipschitz regularization (the last term in the equation), and $\theta_{adv}^i$ is the $i$-th layer's weights of the adversarial policy.

\begin{figure}[t]
  \begin{center}
    \includegraphics[clip,  bb= 0 0 400 490,  width=0.8\columnwidth]{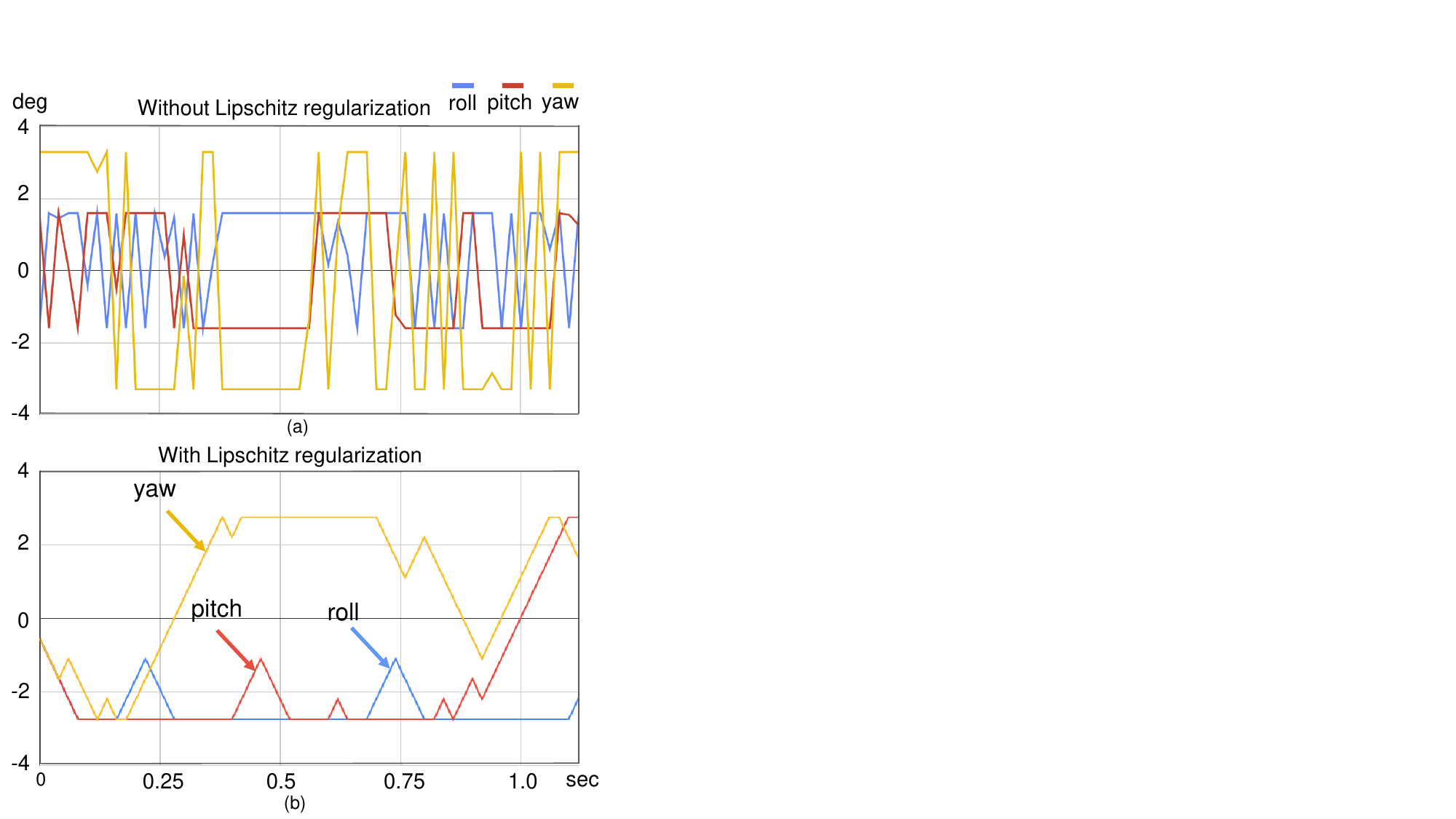}
    \caption{The learned attack sequences under different methods, both able to cause the controller to fail in simulation. (a) is from the vanilla reinforcement learning method, whose adversarial output is in the bang-bang fashion and less practical; (b) is from the proposed method, whose adversarial output is smooth and more realistic for real-world scenarios. As shown in Fig.~\ref{figure:state-estimator-GT}, real-world errors do not frequently oscillate between positive and negative values.}
    \label{figure:lipschitz}
  \end{center}
  \vspace{-4mm}
\end{figure}

RL solutions to our adversarial attack problem empirically adopt a "bang-bang" strategy to maximize the robot's instability (e.g., Fig.~\ref{figure:lipschitz}), but realistic adversarial attacks should be more smooth, like the real-world data in Fig.~\ref{figure:state-estimator-GT}.
Therefore, we regularize the infinity norms of the weights to constrain the Lipschitz constant of the adversary policy network, thereby making the outputs more smooth, inspired by~\cite{shi2019neural} and~\cite{liu2022learning}. Here we assume the adversary policy is a fully-connected network with 1-Lipschitz activation functions (such as ReLU), so the product of infinity norms make an upper bound of the Lipschitz constant~\cite{liu2022learning}. In practice, we use the PPO algorithm~\cite{schulman2017ppo} and add the regularization term to the loss.


The auxiliary rewards are
\begin{equation}
\label{eq:reward}
\begin{split}
     r_t^{aux} = & c_{orient}\cdot g_z + c_{shake}\cdot \left(\omega_x^2+\omega_y^2\right) \\ 
     & + c_{torque}\cdot\sum\nolimits_j {\rm ReLU}(\frac{\lvert\tau_j\rvert}{\tau_{\lim,j}}-1),
\end{split}
\end{equation}
where $g_z$ is the $z$ value of the normalized projected gravity ($-1$ for standing and $1$ for lying), $\omega$ is the angular velocity in the base frame, $\tau$ is the joint torques, and $\tau_{\lim}$ is the soft torque limits. These three terms respectively encourage the bad orientation, the shaking base, and the torques out of limit. They guide the learning of effective adversarial attacks with dense rewards.

\subsection{Close the Loop: Finetuning with Adversarial Samples}

Adversarial attacks not only expose the vulnerabilities of the neural controller, but can also be used for finetuning to further improve the control policy against the weaknesses. Finetuning can be done by simply substituting a portion of the robots' DR with adversarial samples.

\Revise{
Specifically, our finetuning process involves repeating the policy learning with warm-started actor and critic networks based on the previous checkpoint. During this phase, some of the rollouts continue to undergo domain randomization (DR) as in the initial training, while the remainder are subjected to adversarial attacks that are generated based on their states and actions. 
The optimization objective for both training and finetuning of the control policy remains the same, with the only difference being the state and action distributions due to adversarial attacks.
}

Given that adversarial samples can easily make robots fail, applying them to a high proportion of robots can lead to over-conservative behaviors. Therefore, it is important to balance the performance-robustness trade-off during finetuning.


\section{Case Studies}

Our proposed method is not limited by the robot choice and the neural controller design. In this paper, we use the ANYmal robot~\cite{hutter2016anymal} and train two control policies to exemplify our method. The first one is an end-to-end Multi-Layer Perceptron (MLP) policy proposed by \citet{rudin2022corl} for blind locomotion, used as a didactic toy example because of its near-minimal implementation. The second one is the state-of-the-art robust locomotion policy which has gated recurrent units (GRU)~\cite{cho2014learning} in the network, uses central pattern generator (CPG) parameters as the action~\cite{ijspeert2008central}, used in a mission of the DARPA Subterranean (SubT) Challenge~\cite{miki2022scirob, rslsubt2022}.

\subsection{Didactic Example: Adversaries on Minimal Blind Policy}
\label{case:blind}

\begin{table}[t]
\centering
\begin{threeparttable}
\caption{Observations of Attacked Locomotion Policies}
\label{TAB:obs_locopolicy}
\begin{tabular}{l|ll}
\hline
\textbf{Observation type} & \textbf{Input}                                                                                                                                                       & \textbf{Dims}                                                    \\ \hline
Proprioception$^{[B, M]} $           & \begin{tabular}[c]{@{}l@{}}command\\ gravity vector\\ body velocity\\ joint position\\ joint velocity\\ previous joint target\end{tabular}                          & \begin{tabular}[c]{@{}l@{}}3\\ 3\\ 6\\ 12\\ 12\\ 12\end{tabular} \\ \hline
Exteroception$^{[M]}$             & height scan samples                                                                                                                                                  & 208                                                              \\ \hline
History and others$^{[M]}$        & \begin{tabular}[c]{@{}l@{}}joint position history (3 time steps)\\ joint velocity history (2 time steps)\\ joint target history\\ CPG phase information\end{tabular} & \begin{tabular}[c]{@{}l@{}}36\\ 24\\ 12\\ 13\end{tabular}        \\ \hline
\end{tabular}
\begin{tablenotes}[flushleft]
\small
\item Note [\textit{B}] denotes the blind \textit{didactic policy} in Sec.\ref{case:blind}, [\textit{M}] denotes the perceptive \textit{Miki policy} in Sec.\ref{case:darpa}.
\end{tablenotes}
\end{threeparttable}
\vspace{-4mm}
\end{table}

\subsubsection{Locomotion Policy}
We use the open-source implementation of \cite{rudin2022corl} to train the blind locomotion policy (also called the \textit{didactic policy}), which takes only single-frame proprioception as inputs (see also Table~\ref{TAB:obs_locopolicy}). The policy network of the controller is an MLP with $2$ hidden layers, and outputs joint targets to be tracked by a PD controller.

\subsubsection{Adversary Settings}
Regarding the adversarial policy, we use the same observations as that of the locomotion policy, and an MLP with $2$ hidden layers as the policy network. The outputs are the pushing forces on the base in the $x-y$ plane updated at the same frequency of the locomotion policy with the maximum force $100$ N on each axis.

\subsubsection{Attack, robustification, and reattack}
\label{subsubsec:arr4stage}
\begin{figure}[t]
  \begin{center}
    \includegraphics[clip,  bb= 0 0 650 362,  width=1.0\columnwidth]{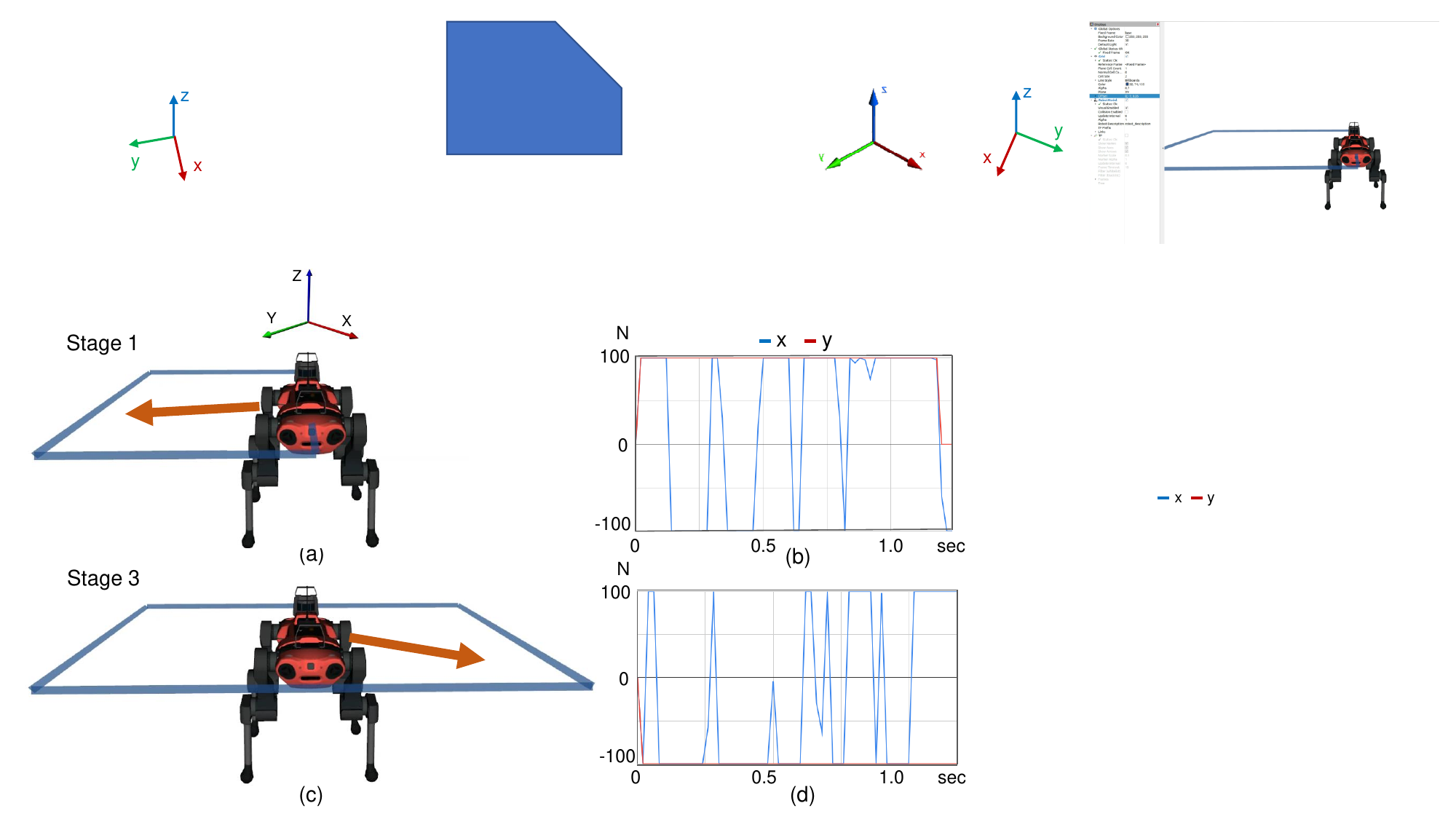}
    \caption{Didactic example to show the proposed method being effective. (a)-(b): Stage 1: we limit the $y$ values of the adversarial forces to be positive (c)-(d): Stage 3: we allow the $y$ values of the adversarial forces to be either positive or negative when attacking the finetuned controller. During reattacking, the learned adversary effectively finds the unpatched weakness by leveraging the previously unused negative $y$ forces.}
    \label{figure:case-study-1}
  \end{center}
  \vspace{-4mm}
\end{figure}

To show that our adversarial attacks can effectively cause failures, we conduct a simple controlled experiment in simulation:
\begin{enumerate}
\setcounter{enumi}{-1}
    \item We command the robot to walk forward on the flat terrain with $0.4$ m/s.
    \item In Stage 1, we purposely set the adversarial force to have only $x$ and positive $y$ values as Fig.~\ref{figure:case-study-1}(a). The learned adversarial attacks can make the robot fall over.
    \item In Stage 2, we finetune the original control policy with the learned adversarial attacks in Stage 1 to improve its robustness against the biased perturbations.
    \item In Stage 3, we train a new adversary to attack the robustified policy, and allow the force to have both positive and negative $y$ values as Fig.~\ref{figure:case-study-1}(c). The attack succeeds again and tends to leverage perturbations from the negative $y$ axis.
\end{enumerate}
This case presents a basic pipeline of attack-defense-reattack, and will be reused for further discussion in Sec.~\ref{sec:analyses}.

\subsubsection{Baselines}
To demonstrate the necessity and efficacy of the learning-based adversarial attack, we also conduct experiments for the following standard test (ST) settings as baselines:
\begin{itemize}
    \item ST 1: apply a random constant pushing force~\cite{lee2020scirob};
    \item ST 2: apply random pushing forces at 0.5 Hz~\cite{weng2023comparability};
    \item ST 3: apply maximum pushing forces in random directions at 0.5 Hz;
    \item ST 4: apply a random impact~\cite{zhangei2023quadruped-tro, weng2023standard-legged-test}, here we adjust it by applying the perturbation force for 0.2 s to ensure a controlled comparison.
\end{itemize}
The attack range remains consistent with Stage 1, where perturbations are applied to the robot's base, utilizing the \textit{didactic policy} as the locomotion controller. 

Our simulation experiments show that, \textit{none} of the four STs above can induce any failure through $1000$ trials, whereas the learned adversary can fail the controller with a $100\%$ probability.
Although these STs are useful for evaluating the effects of large perturbations, they prove incapable of uncovering the subtle perturbations that are the focus of our study.

\subsection{Attacks on DARPA-SubT-Winning Locomotion Policy}
\label{case:darpa}

\subsubsection{Locomotion Policy}
In this case, the locomotion policy is a robust perceptive policy proposed by \citet{miki2022scirob} (also called the \textit{Miki policy} in this paper) trained for challenging scenarios like subterranean caves and snowy mountains. The policy is empirically robust against observation noises and external perturbations~\cite{rslsubt2022}, and the failure cases are rare in the real world. 

\subsubsection{Adversary Settings}
Regarding the adversary, the policy network is an MLP with $3$ hidden layers, and the observations are the same as that of the locomotion policy (see Table~\ref{TAB:obs_locopolicy}). The attacks can be applied to the observation space, the command space, and the perturbation space, updated at the same frequency of the locomotion policy. The attacked values and their ranges are listed in Table \ref{TAB:adv_range_darpa}, with the change rate limited to $0.1\times$~maximum value per timestep (0.02 s).

In the following experimental results, we employ two settings. 
One (Sec.~\ref{subsubsec:sim2real_adv}) attacks only the observation space and the command space to verify consistency between simulation and the real world, given that perturbations cannot be accurately replicated in the real world.
The other (Sec.~\ref{subsubsec:robustify_darpa}) attacks all of the three spaces, and the learned adversary is used in simulation to further robustify the locomotion policy.

\begin{table}[t]
\centering
\caption{Adversarial Space and Range Values against \textit{Miki Policy}}
\label{TAB:adv_range_darpa}
\begin{tabular}{l|l}
\hline
Adversarial space  & \multicolumn{1}{l}{Range values}                                                      \\ \hline
Twist command   & \begin{tabular}[c]{@{}l@{}}x/y: $(-0.5, 0.5)$ m/s\\ yaw: $(-0.5, 0.5)$ rad/s\end{tabular} \\ \hline
Orientation error     & Roll/pitch/yaw: $(-3.0^{\circ}, 3.0^{\circ})$                                                           \\ \hline
Perturbation force & End effectors: $(-15.0, 15.0)$ N                                                          \\ \hline
\end{tabular}
\end{table}

\subsubsection{Learned Adversary with Real-World Validation}
\label{subsubsec:sim2real_adv}
Our method can effectively learn adversarial attacks to fail the locomotion policy with only the observation space and the command space.
To verify that the simulation results align with the real-world performance, we conduct the same attacking test in simulation and the real world. We also try rescaled orientation attacks to identify the minimal range to cause the failure. 
The motion of the robot under attack is depicted in Fig.~\ref{figure:fall}.

\begin{figure}[t]
  \begin{center}
    \includegraphics[clip,  bb= 0 0 250 380,  width=0.75\columnwidth]{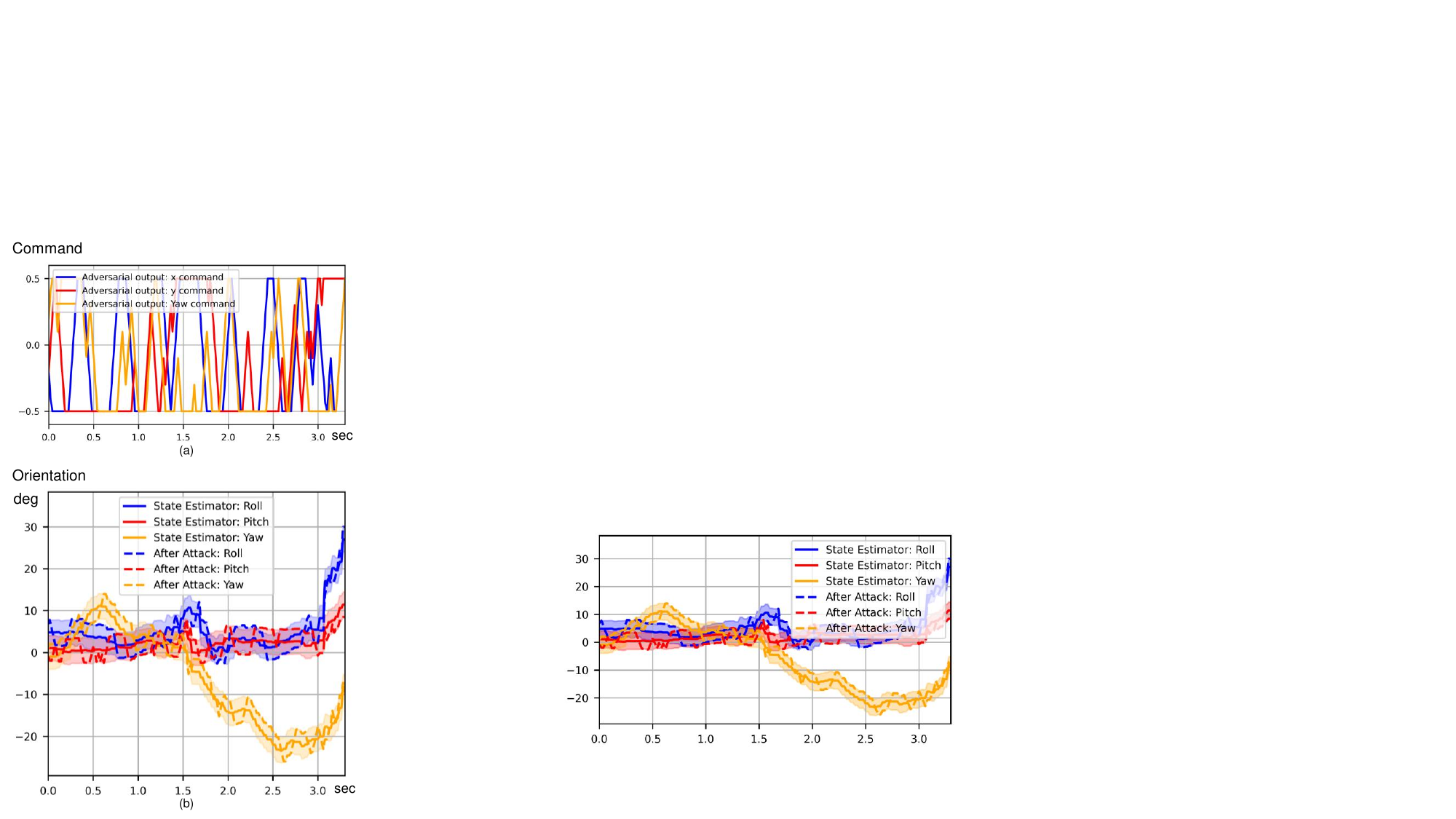}
    \caption{Plots of adversarial output to fall over the real robot. (a) is the adversarial attack sequence in the command space; (b) is the attack sequence in the observation space, which outputs the orientation errors within $3^{\circ}$. Note that to avoid the damage on the physical hardware, early-stop is triggered before the robot falls down. }
    \label{figure:real-adversarial-cmd-euler}
  \end{center}
  \vspace{-3mm}
\end{figure}

\begin{figure}[t]
  \begin{center}
    \includegraphics[clip,  bb= 0 0 380 180,  width=1.0\columnwidth]{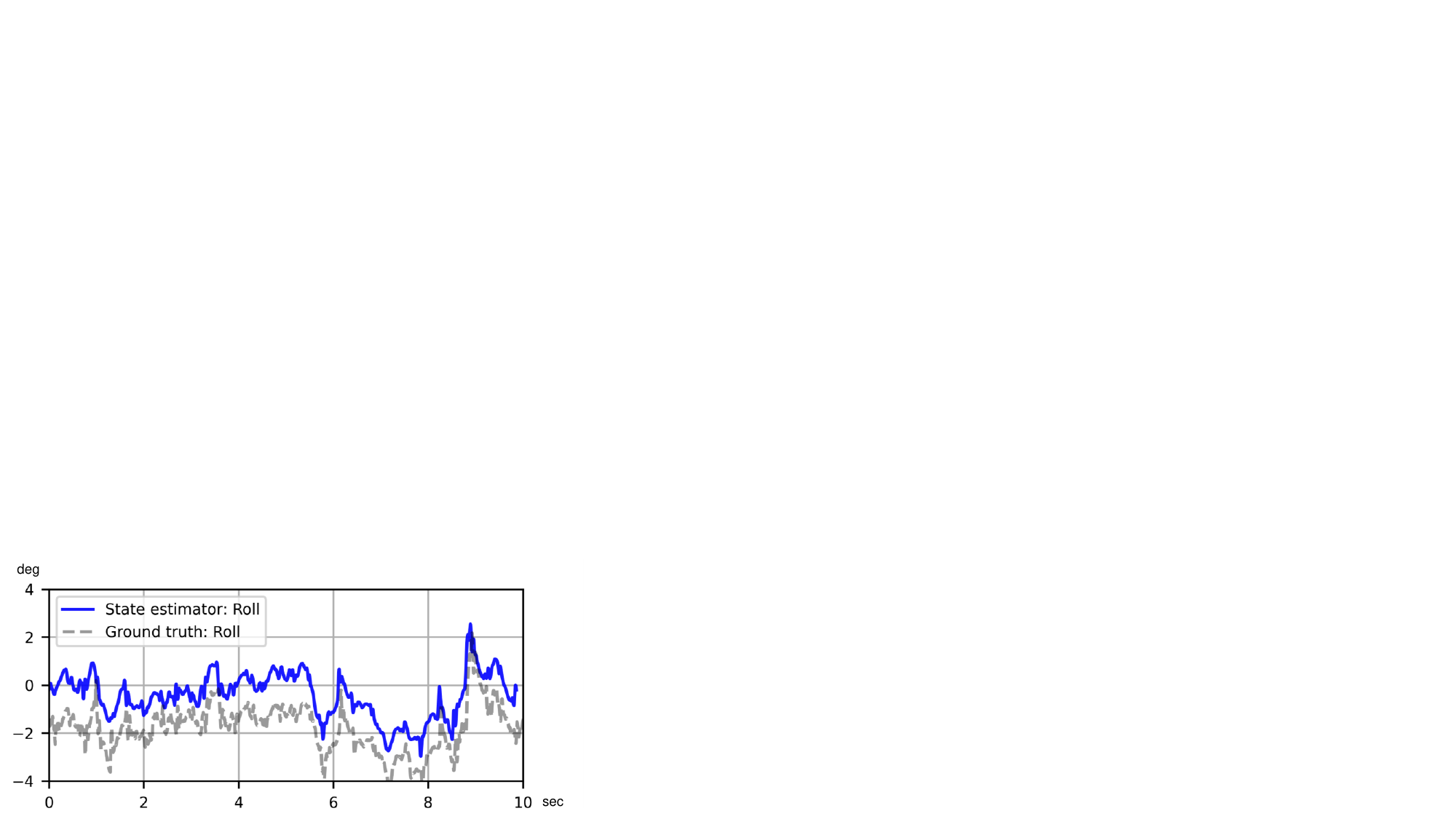}
    \caption{Orientation data from the state estimator and ground truth (via motion capture) when the robot traverses flat indoor terrain, with peak errors reaching almost $3^{\circ}$. These estimation errors tend to significantly increase on uneven or slippery outdoor surfaces, or when the robot is subjected to perturbations~\cite{bloesch2013state}.}
    \label{figure:state-estimator-GT}
  \end{center}
  \vspace{-3mm}
\end{figure}

\begin{figure}[t]
  \begin{center}
    \includegraphics[clip,  bb= 0 0 550 540,  width=1.0\columnwidth]{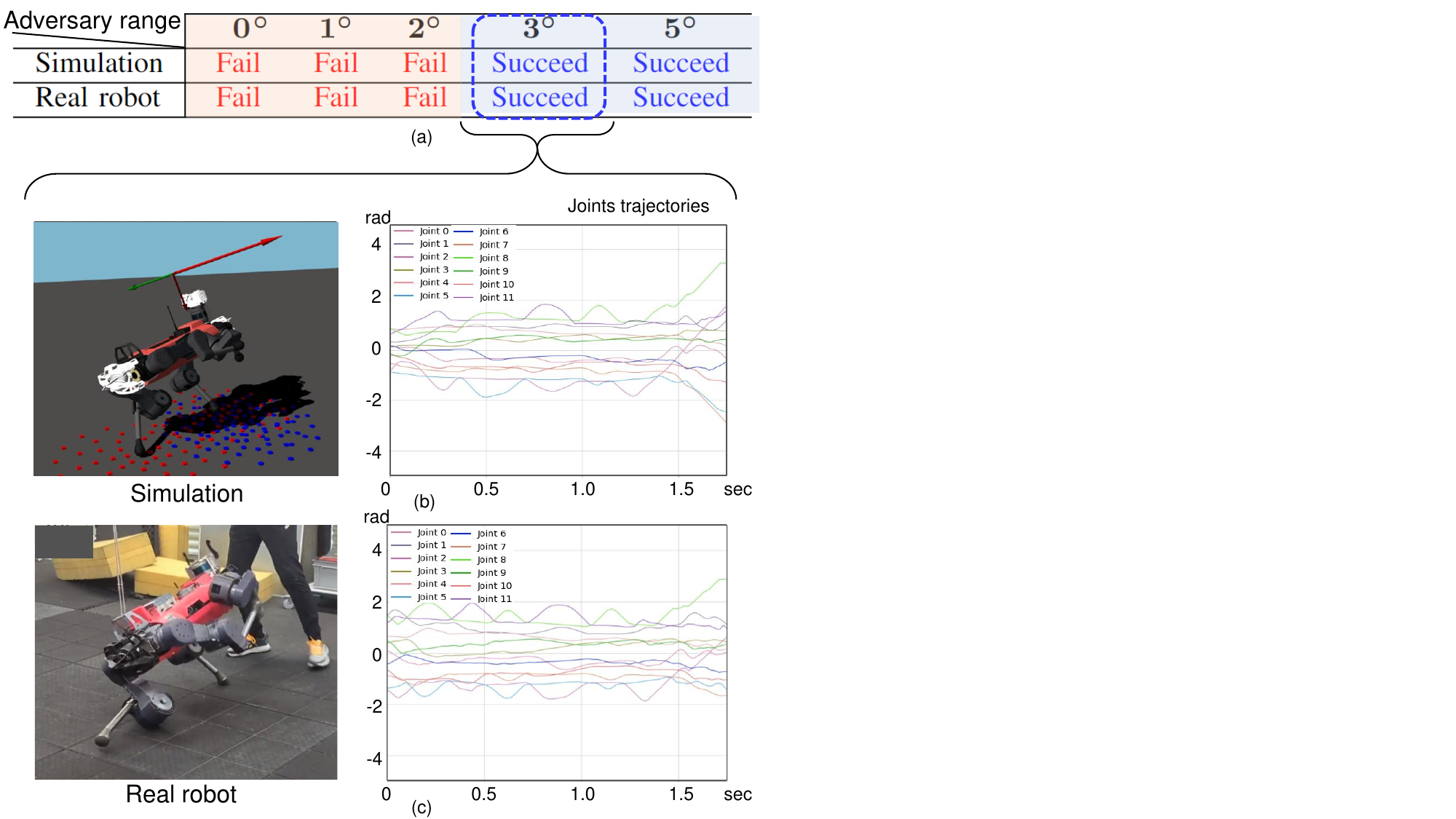}
    \caption{Sim-to-real comparison under adversarial attacks. (a) shows the adversary cannot successfully attack the controller when the orientation error is rescaled to be no larger than $2^\circ$. This applies to both simulation and the real robot. (b) is falling-over snapshots and the joint trajectories in simulation when the orientation error is bounded by $3^\circ$. (c) is falling-over snapshots and the joint trajectories in the real world when the orientation error is bounded by $3^\circ$.}
    \label{figure:sim2real-mismatch}
  \end{center}
  \vspace{-3mm}
\end{figure}

The test results presented in Fig.~\ref{figure:sim2real-mismatch} demonstrate the sim-to-real transferability of our learned adversary, and the corresponding attack sequence is visualized in Fig.~\ref{figure:real-adversarial-cmd-euler} with comparison to a typical state estimation error curve in Fig.~\ref{figure:state-estimator-GT}.
These results indicate that, vulnerabilities identified in simulation do reflect the controller's weaknesses in the real world, and the risk of encountering such failures during operation is non-negligible.

\subsubsection{Robustification}
\label{subsubsec:robustify_darpa}
\begin{figure}[t]
  \begin{center}
    \includegraphics[clip,  bb= 0 0 720 410,  width=1.0\columnwidth]{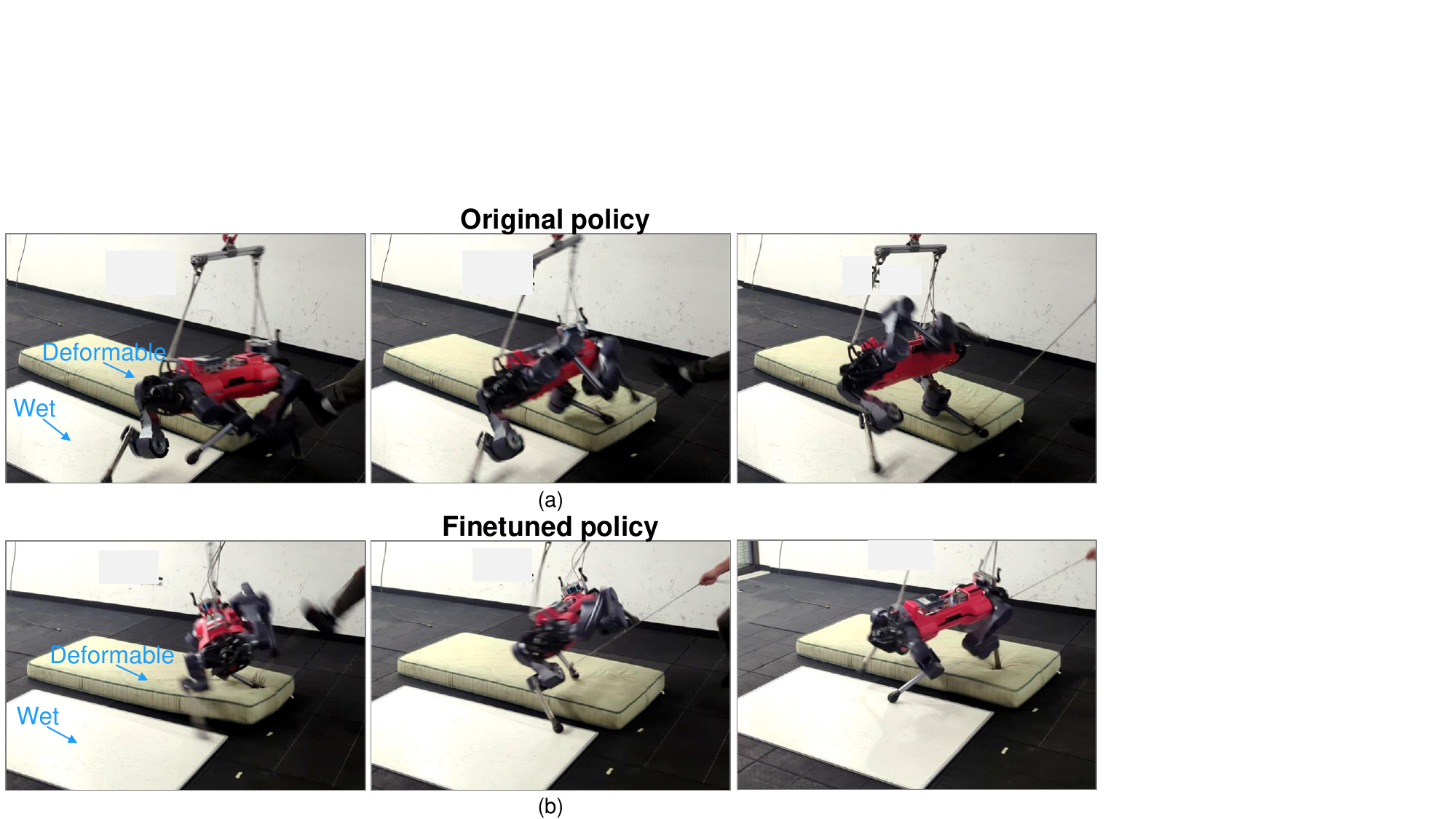}
    \caption{Challenging indoor perturbation experiments on on a slippery wet whiteboard and a deformable mattress. (a) shows the performance of the original \textit{Miki policy}, which quickly loses balance; (b) is the robustified policy by our proposed method, showcasing more robust reactive behaviors.
    }
    \label{figure:indoor-taka-finetne-policy-compare.pdf}
  \end{center}
\end{figure}

\begin{figure}[t]
  \begin{center}
    \includegraphics[clip,  bb= 0 0 650 520,  width=1.0\columnwidth]{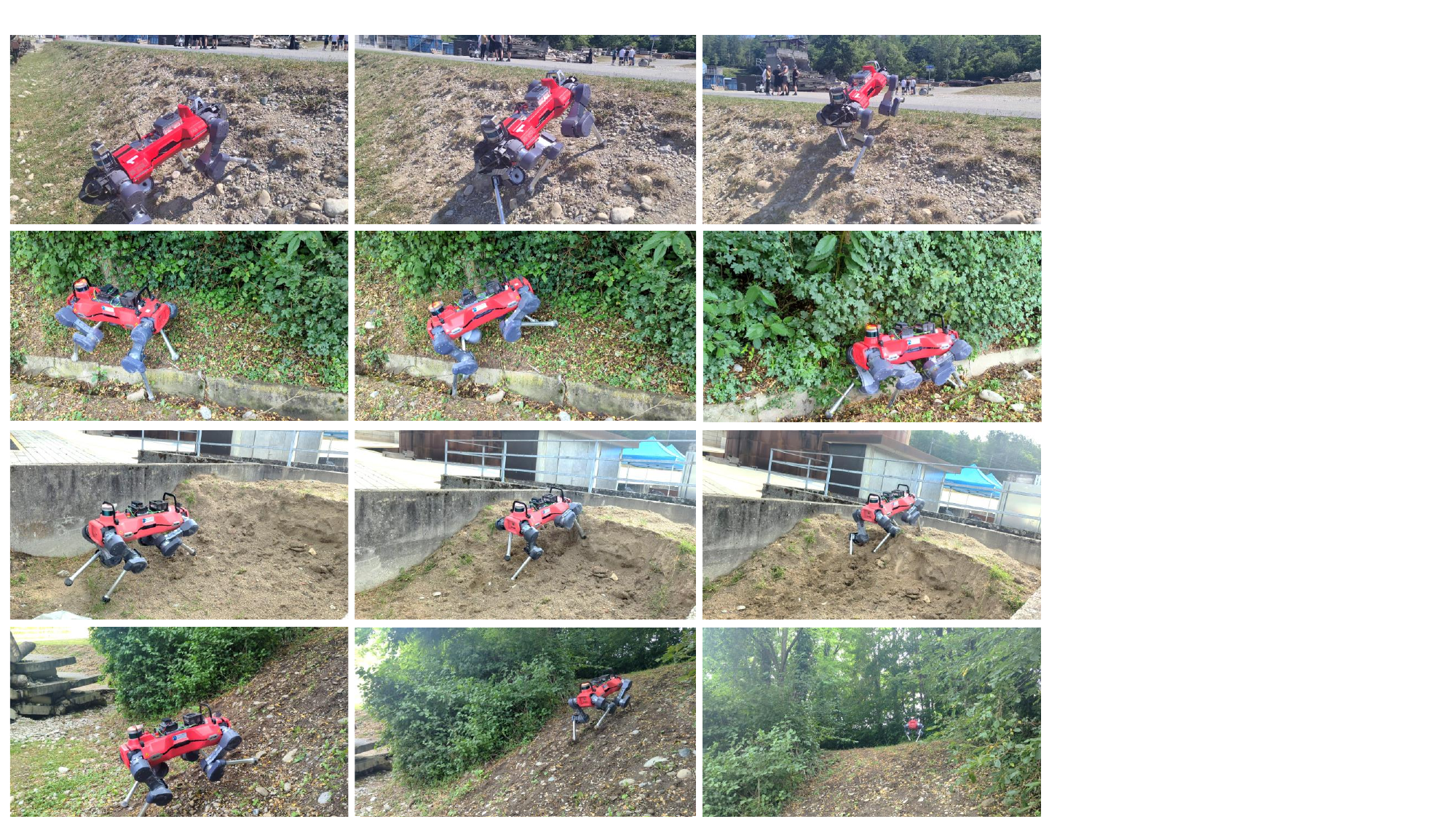}
    \caption{Outdoor experiments in the challenging terrains, including rough slopes, slippery ditches, deformable sand, and steep mountain roads with small rocks. The finetuned policy maintains the traversability while being more robust against adversarial scenarios.}
    \label{figure:outdoor-snapshots}
  \end{center}
  \vspace{-3mm}
\end{figure}

We train an adversary with all of the three adversarial spaces, and finetune the policy in simulation with a $5\%$ probability of encountering the learned adversary instead of randomized perturbations. We expect that the finetuned policy is more robust against perturbations and state estimation errors. This is verified not only by a failed trial of reattack learning, but also by the real-world robust performances indoors (Fig.~\ref{figure:indoor-taka-finetne-policy-compare.pdf}) and outdoors (Fig.~\ref{figure:outdoor-snapshots}).

\begin{figure}[t]
  \begin{center}
    \includegraphics[clip,  bb= 0 0 570 450,  width=1.0\columnwidth]{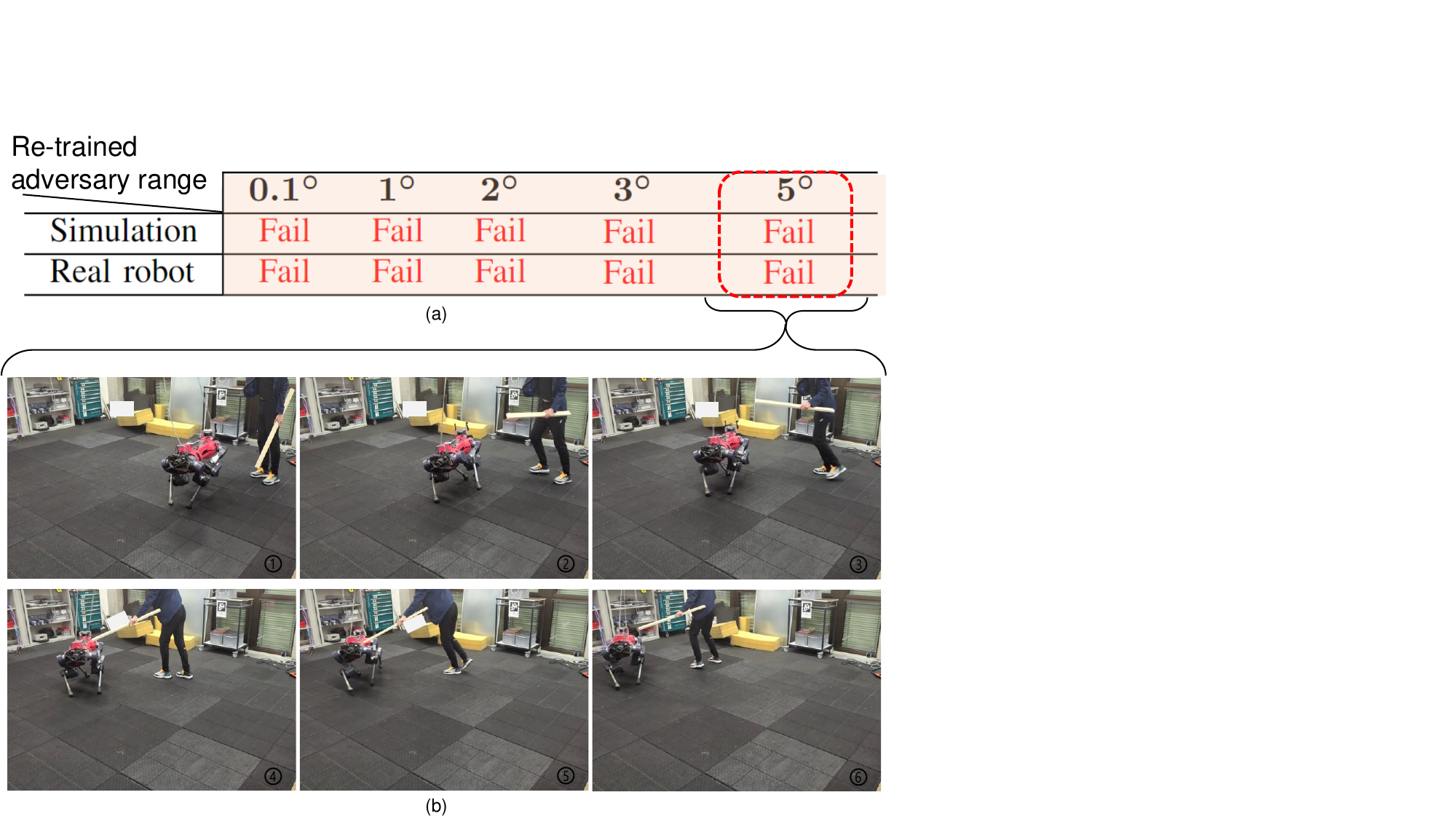}
    \caption{\Revise{Re-trained adversaries fail to attack the finetuned controller. (a) Compared to the controller before finetuning in Fig.~\ref{figure:sim2real-mismatch}, the finetuned controller can withstand adversaries at a larger scale. The orientation error is bounded by $5^\circ$, informed by expert knowledge for realistic attack scenarios. (b) In real-world conditions, the finetuned controller maintains its robustness, effectively resisting both re-attacks and additional perturbations applied by the operator.}}
    \label{figure:fine-tune-reattack}
  \end{center}
  \vspace{-3mm}
\end{figure}

\Revise{
To further assess the robustness of the finetuned policy, we conduct a re-attack at a larger scale, , with the results presented in Fig.~\ref{figure:fine-tune-reattack}. During the real-world test, the operator also applied additional perturbations. Despite these challenging conditions, the robot maintained its robustness and resisted all attacks.
}

To quantitatively examine the performance change of the new controller, we compare the joint torques and command tracking accuracy before and after robustification, as presented in Table~\ref{table:finetune_policy_performance}. Despite the improved robustness, we find no significant change in the tracking performance.

\begin{table}[t]
\centering
\caption{Comparison of Torques and Tracking Performance Before and After Robustification of \textit{Miki Policy} in Different Terrains}
\label{table:finetune_policy_performance}
\begin{tabular}{c|cc}
\hline
\textbf{Terrain} & \textbf{Avg. joint torque} & \textbf{Avg. tracking error} \\ \hline
Flat terrain     & $+ 0.42$ Nm                      & $+ 0.002$ m/s                   \\ \hline
Rough terrain          & $- 1.76$ Nm                      & $- 0.006$ m/s                  \\ \hline
Standard stairs           & $+ 0.26$ Nm                      & $+ 0.008$ m/s                  \\ \hline
Random stones          & $- 0.26$ Nm                     & $+ 0.006$ m/s                  \\ \hline
\end{tabular}
\vspace{-3mm}
\end{table}

\section{Analyses and Extensive Studies}
\label{sec:analyses}

\subsection{Is DR sufficient for controller robustness?}
\label{sec:extensive-dr-not-safe}

Domain randomization (DR) randomizes the properties of the environments, with the expectation that the policy will work across all these varied settings. 
However, we \textit{hypothesize} that naive randomization methods (e.g., time-invariant uniform distributions or Gaussian distributions) may hardly cover threatening time-sequential adversaries in the high-dimensional adversarial space. Consequently, DR may be insufficient in guarding the controller against adversarial attacks.

To verify this, we train a locomotion policy similar to the \textit{didactic policy} and with exteroception observations. During training, we add randomized perturbation forces up to $100$ N and varies at $5$ Hz. We then train an adversary policy with up to $100$ N perturbation forces to attack this policy. As shown by Fig.~\ref{figure:safe-set} ("policy-DR"), the robot still fails under the learned adversarial attacks, though it can survive larger pushing forces compared to the policy without DR.

Our conclusions are: 1) DR is not sufficient to ensure the robustness against adversarial attacks; meanwhile, 2) adversarial attacks can be complementary to DR.

\subsection{Attacks as robustness indicators}

\begin{figure}[t]
  \begin{center}
    \includegraphics[clip,  bb= 0 0 450 340,  width=0.9\columnwidth]{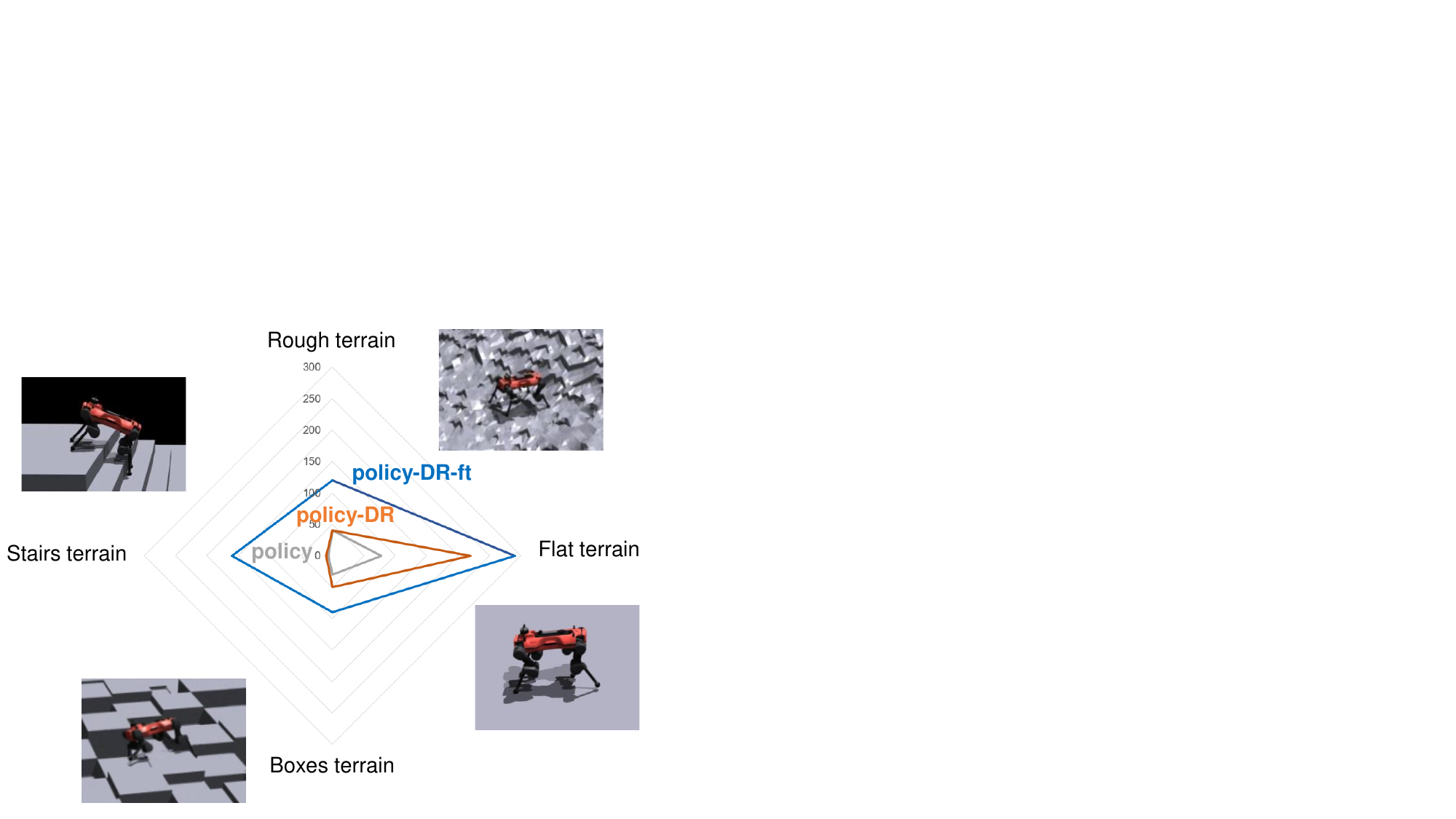}
    \caption{\Revise{Minimum perturbation force ranges that cause the controller to fall over vary across different terrains, highlighting that finetuning with learned adversarial attacks can significantly diminish its vulnerabilities on challenging, uneven terrains.}}
    \label{figure:safe-set}
  \end{center}
\end{figure}

Quantitatively assessing the robustness of control policies remains a challenge, as standard tests are inefficient in identifying worst cases. 
As an initial exploration, we claim that 
adversarial attacks, as the effective tools to search the worst cases, can be leveraged to indicate the robustness of a control policy.

we conduct the following experiments to demonstrate this: 
\begin{enumerate}
    \item We train a locomotion policy with the same random perturbation setting as \ref{sec:extensive-dr-not-safe} (called "policy-DR"). We train another one without DR (called "policy").
    \item We add the perturbation ranges and exteroception to the adversary's observation space. We also randomize the perturbation ranges during training so that the agent is aware of the maximum perturbation magnitude.
    \item We train adversaries to respectively attack the two locomotion policies.  We then finetune "policy-DR" under adversaries. This refined version of the policy, now termed "policy-DR-ft", is then re-examined under newly learned adversarial attacks.
    \item We execute the three adversarial policies on the corresponding locomotion policies. During execution for each, we gradually reduce the perturbation range until the locomotion policy can withstand its adversary.
    \item We record the minimal ranges to make the control policies fail in different terrains. The values are reported in Fig. \ref{figure:safe-set}.
\end{enumerate}
As we can see, Fig. \ref{figure:safe-set} clearly shows  "policy-DR" is better than "policy", while "policy-DR-ft" shows the best safety. This example shows how adversarial attacks can be potentially used to assess the robustness of neural controllers.

\subsection{Multi-modality is essential}

Previous works in legged robots only investigate a single adversarial modality \cite{zhangei2023quadruped-tro, weng2023standard-legged-test, weng2023comparability, tanjie2020rarl}.
Here we claim that, \textit{multi-modal} adversaries are more effective in searching for the worst cases.

Using the \textit{Miki policy} in Sec.~\ref{case:darpa} as the example, we train adversaries with different adversarial spaces, and the results are presented in Table~\ref{TAB:adv-space-compare}.

\begin{table}[t]
\centering
\caption{Comparison of Different Combinations in Adversarial Space }
\label{TAB:adv-space-compare}
\begin{tabular}{c|cccllll}
\hline
\textbf{}           & \textbf{C} & \textbf{O} & \textbf{P} & \textbf{C+O} & \textbf{C+P} & \textbf{O+P} & \textbf{C+O+P} \\ \hline
Adversaries results & \color{red}{F}       & \color{red}{F}       & \color{red}{F}       & \multicolumn{1}{c}{\color{blue}{S}}      & \multicolumn{1}{c}{\color{blue}{S}}      & \multicolumn{1}{c}{\color{blue}{S}}      & \multicolumn{1}{c}{\color{blue}{S}}        \\ \hline
\end{tabular}
\begin{tablenotes}[flushleft]
\small
\item \textbf{C/O/P} denotes the command/observation/perturbation space, S/F denotes the success/failure to cause robots to fall over with the corresponding adversarial space.
\end{tablenotes}
\vspace{-2mm}
\end{table}

Based on the results, we conclude that multiple modalities are essential in finding effective and subtle adversaries. 
Combining any two or all three of the adversarial subspaces can achieve successful attacks despite larger search spaces, while no unimodal attacks can succeed as we limit the attack ranges.
Multi-modal adversaries also better reflect the real world where failures of a well-developed controller are often due to the combination of multiple factors.

\Revise{
\subsection{Ablating reward function design}

A conventional approach in adversarial reinforcement learning sets the adversarial reward as the negative of the policy reward, thus formulating the problem within a zero-sum context. However, the complexity of various robot control challenges motivates the development of distinct policy and adversarial rewards \cite{peng2021amp, peng2022ase, marko2023adv-motor-prior, li2023learning, li2023versatile}, raising the question of which approach is more effective for our problem addressing complex locomotion reward structures in real-world applications. Furthermore, we aim to validate the necessity of our auxiliary terms as delineated in Equations \ref{eq:rl-adv} and \ref{eq:reward}.

To illustrate the need for distinct adversarial rewards, we conducted an ablation study. Initially, we adopted the zero-sum formulation, setting the adversarial reward as the negative of the original policy's reward. This configuration was ineffective for achieving our objective of inducing the robot to fall over. To more comprehensively evaluate the effectiveness of our proposed adversarial reward structure, we conducted further ablation analyses by systematically removing each component; the outcomes are detailed in Table~\ref{TAB:adv-reward-compare}. Additionally, the auxiliary reward facilitated greater exploration in the adversarial space, enabling the generation of diverse failure scenarios \cite{croce2020ensemble_adv, liu2021multi}. More details are provided in Section~\ref{subsec:diverse}.


We conclude that in real-world applications, many components of the original control policy are specifically designed to shape complex behaviors, including metrics such as velocity tracking, foot clearance, and torque penalties. Simply inverting the policy reward directs the adversary towards minimizing task performance (evidenced by poor velocity tracking and enlarged torques) rather than inducing catastrophic failures. This inefficacy arises in our problem because a fall produces a reward of $0$, significantly lower than the penalties from suboptimal walking. Similar challenges are observed in systems using cross-entropy loss \cite{croce2020ensemble_adv}. While recent theoretical advances have begun to more thoroughly investigate non-zero-sum adversarial training, employing multi-objective optimization \cite{albuquerque2019multi} or integrating the adversarial objective as a constraint \cite{robey2023nonzerosum}, the domain still demands extensive exploration to fully develop and clarify the underlying theories.

\begin{table}[t]
\centering
\caption{\Revise{Ablative Analysis of Various Adversarial Reward Combinations}}
\label{TAB:adv-reward-compare}
\begin{tabular}{c|c|c}
\hline
\textbf{\Revise{Adversarial reward setting}}                      & \textbf{\Revise{Convergence iter.}}         & \textbf{\Revise{Falling over}} \\ \hline
\textbf{\Revise{Eq. \ref{eq:rl-adv} (Proposed)}}              & \Revise{240}      & \Revise{Succeed}                         \\ \hline
\textbf{\Revise{Zero-sum fashion}}              & \Revise{320}       & \Revise{Fail}                        \\ \hline
\textbf{\Revise{Set $\mathds{1}_{alive} = 0$}}          & \Revise{280}      & \Revise{Fail}                         \\ \hline
\textbf{\Revise{Set $c_{orien}/c_{shake} = 0$}} & \Revise{650}            & \Revise{Succeed}                    \\ \hline
\textbf{\Revise{Set $c_{torque} = 0$}}         & \Revise{720}            & \Revise{Succeed}                    \\ \hline
\end{tabular}
\begin{tablenotes}[flushleft]
\small
\item \Revise{The first setting is our proposed adversarial reward; in the zero-sum setting, the adversarial reward is set to be the negative of the locomotion reward in \cite{rudin2022corl}; for the other approaches, parts of the auxiliary rewards in Eq.~\ref{eq:rl-adv} and Eq.~\ref{eq:reward} are set to $0$.}
\end{tablenotes}
\vspace{-2mm}
\end{table}


}

\subsection{Can humans be good hackers?}

To improve robustness, control policies are often iteratively refined by humans through intuitive testing of perturbations and sensor noise on the robot. The question arises: can humans efficiently identify the controller's weaknesses, and is there a real need for the computational method proposed in this paper?

To verify this, despite the difficulty in formally evaluating human performance (e.g., inviting a human champion against machine intelligence ~\cite{silver2016mastering} \cite{kaufmann2023champion}), we make a first attempt by organizing a small \textit{AI safety challenge} on locomotion controller hacking.

In the challenge, we provide the \textit{didactic policy} in simulation and ask participants to attack the robot. 
The robot is consistently commanded to walk on flat ground with a fixed forward velocity of $0.4$~m/s.
Participants are allowed to use joysticks to apply 2D pushing forces (up to $100$~N on $x, y$ axes) to the robot base and to overwrite the lateral and yaw velocity commands respectively within $0.5$~m/s and $0.5$ rad/s.

Regarding the machine intelligence counterpart, for fair comparison, we train an adversarial agent without joint state observations as humans have only limited observations visually displayed in the simulator's GUI. Besides, we limit the adversary's output frequency to 5~Hz as humans cannot move the joysticks at high frequencies.

We successfully invited $100$ volunteer participants with a background in robotics to join the challenge. Due to resource constraints, we gave each participant $5$ minutes to attempt their attacks after they learned how to operate the system. During the whole challenge, only $3$ of the participants successfully made the robot fall. We compare the performances of the successful participants and our learned adversary in Table~\ref{TAB:human-hacker}.
Compared to humans, our learned adversary can effectively attack the controller in \textit{less time} and cause more severe damage through \textit{more violent falls}.

\begin{table}[t]
\centering
\caption{Comparison of successful human hackers and proposed adversarial method}
\label{TAB:human-hacker}
\resizebox{1\linewidth}{!}{
\begin{tabular}{l|cccc}
\hline
\multicolumn{1}{l|}{\textbf{}}                                                                & \textbf{Winner 1} & \textbf{Winner 2} & \textbf{Winner 3} & \textbf{Our Adversary} \\ \hline
\textbf{\begin{tabular}[l]{@{}l@{}}Avg. survival\\ time (s)\end{tabular}}                   & $4.0$                & $5.8$               &  $11.5$             & \textcolor{blue}{$1.38$}                       \\ \hline
\textbf{\begin{tabular}[l]{@{}l@{}}Avg. falling\\ roll rate (rad/s)\end{tabular}} & $1.77$              & $1.53$                & $2.80$               & \textcolor{red}{$5.04$}                        \\ \hline
\end{tabular}}
\end{table}

We conclude that, identifying weaknesses in RL-based locomotion controllers through human experience is challenging, even when there is no perturbation randomization for robustness during training. On the other hand, the computational method proposed in this paper is crucial to be a complement and help uncover vulnerabilities in black-box neural controllers.

\begin{figure}[t]
  \begin{center}
    \includegraphics[clip,  bb= 0 0 500 190,  width=1.0\columnwidth]{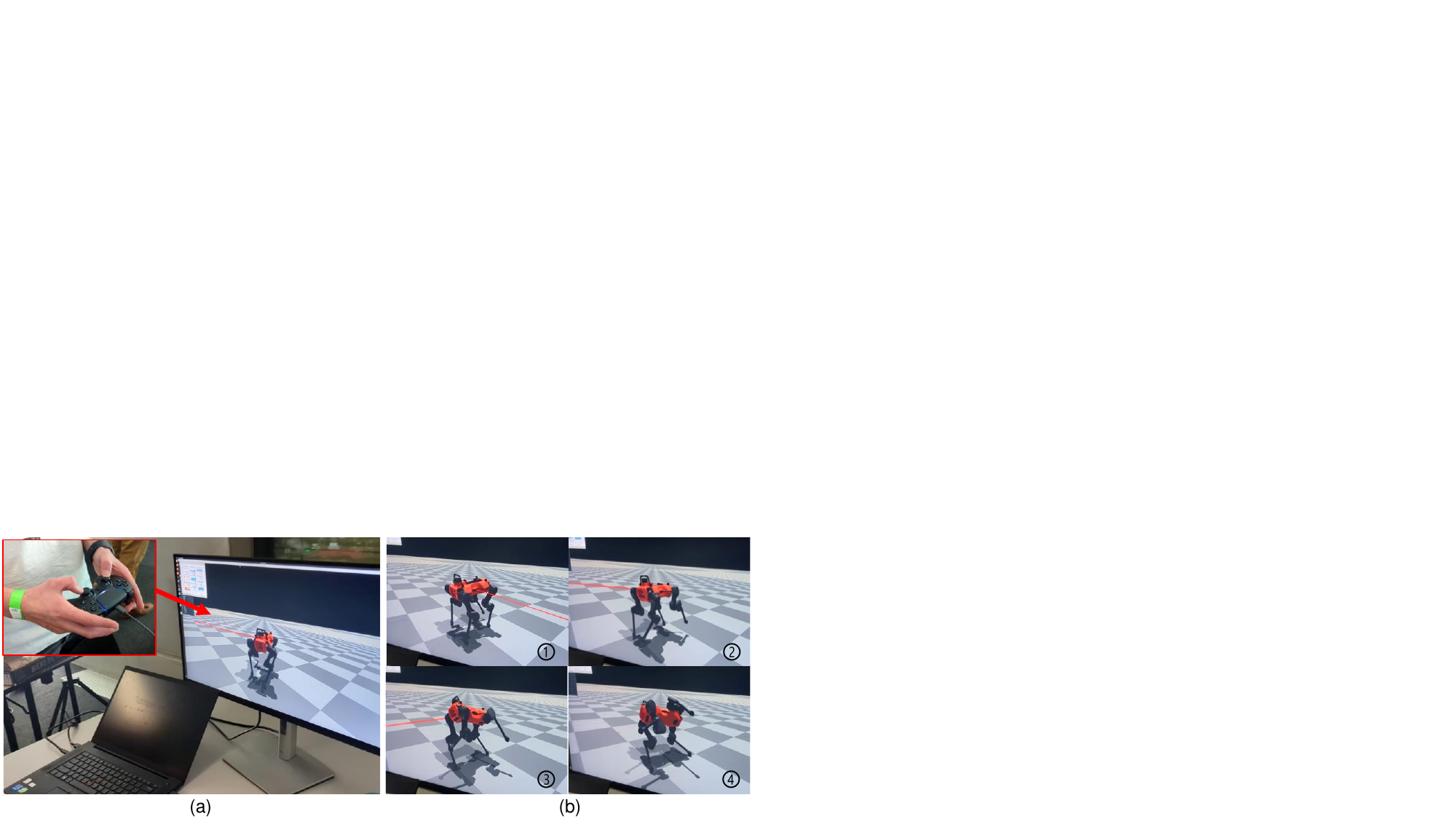}
    \caption{\textit{AI safety challenge} in hacking robot controller. (a): Human can use remote controller to apply perturbation force and send twist command to the robot in the simulation; (b): Snapshots of the Winner 2 in falling over the controller.}
    \label{figure:human-hack}
  \end{center}
  \vspace{-3mm}
\end{figure}

\subsection{Attacks that leverage terrains}

Although our real-world attack experiments are conducted on flat ground for safety reasons, uneven terrains could potentially present more challenging scenarios.

\begin{figure}[t]
  \begin{center}
    \includegraphics[clip,  bb= 0 0 500 540,  width=1.0\columnwidth]{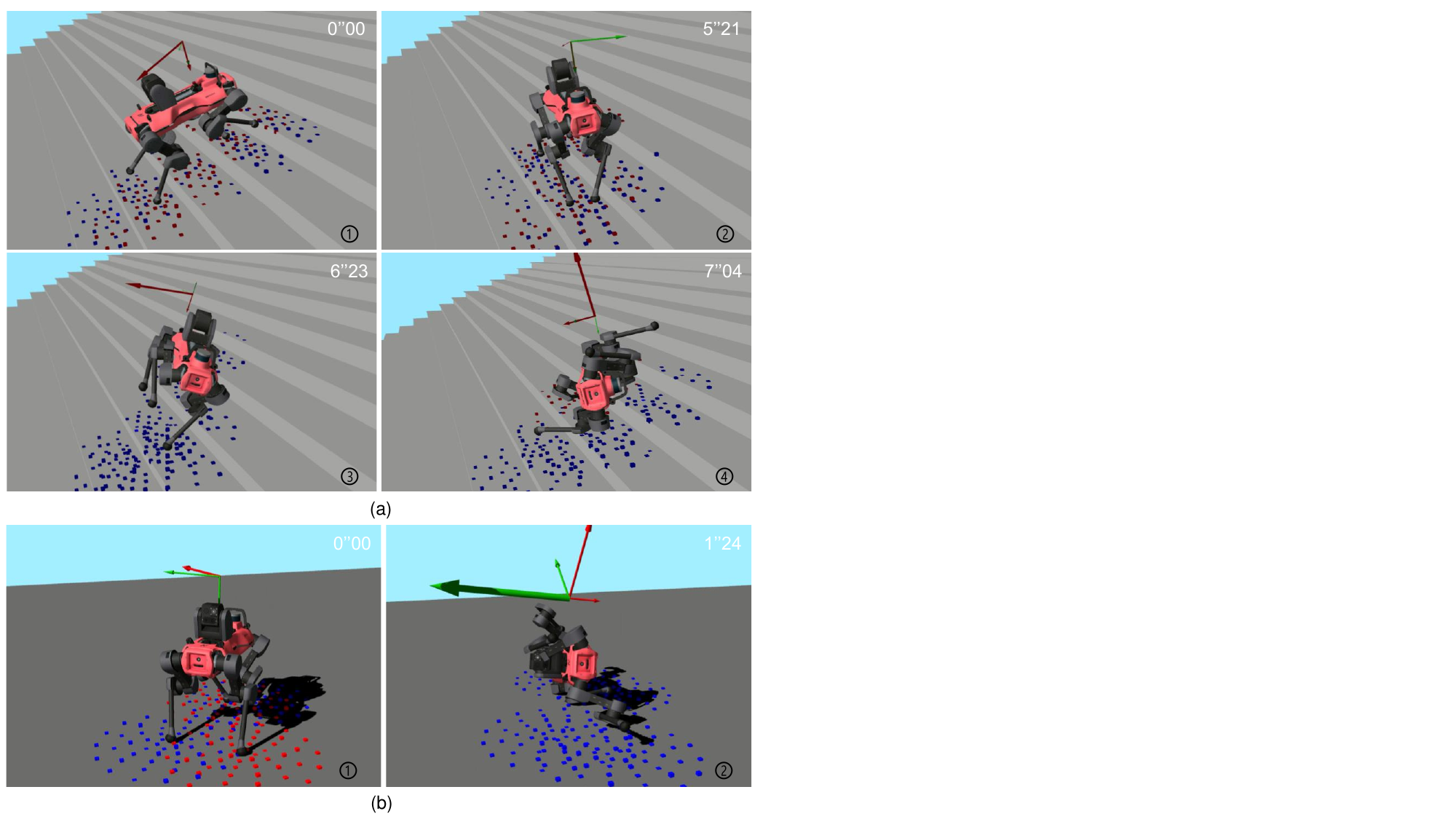}
    \caption{Learned adversarial agent with exteroceptive observations leverages the terrain for effective attacks. (a): Stairs are used strategically, where the adversary first rotates the robot to position it advantageously before initiating a successful fall-over attack. (b): On flat terrain, conversely, the adversary launches the attack directly without the rotation maneuver observed on the stairs.}
    \label{figure:terrain-aware-strategy}
  \end{center}
  \vspace{-3mm}
\end{figure}

Indeed, we find in simulation that the learned adversary against the \textit{Miki policy} in Sec.~\ref{case:darpa}, with exteroceptive observations, can actively exploit uneven terrains to enhance the attack. Fig.~\ref{figure:terrain-aware-strategy}(a) illustrates an instance where the learned adversary first directs the robot towards the edge of stairs and then utilizes the stairs to induce a fall. In contrast, such behavior is not observed on flat ground, as depicted in Fig.~\ref{figure:terrain-aware-strategy}(b).

\subsection{Can we uncover diverse weaknesses?}
\label{subsec:diverse}

The proposed RL-based method empirically converges to a specific attack strategy in each run, while the controller's weaknesses may be non-unique. 
Hence, it is preferable to uncover a variety of weaknesses rather than a single one.

The problem can be addressed in two ways. One is to train different adversaries with different random seeds for neural network initialization~\cite{zhang2020diversified}, which lacks controllability. The other is to guide the adversaries with diverse reward settings~\cite{yiwu2020randreward} as shown here.

\begin{figure}[t]
  \begin{center}
    \includegraphics[clip,  bb= 0 0 590 160,  width=1.0\columnwidth]{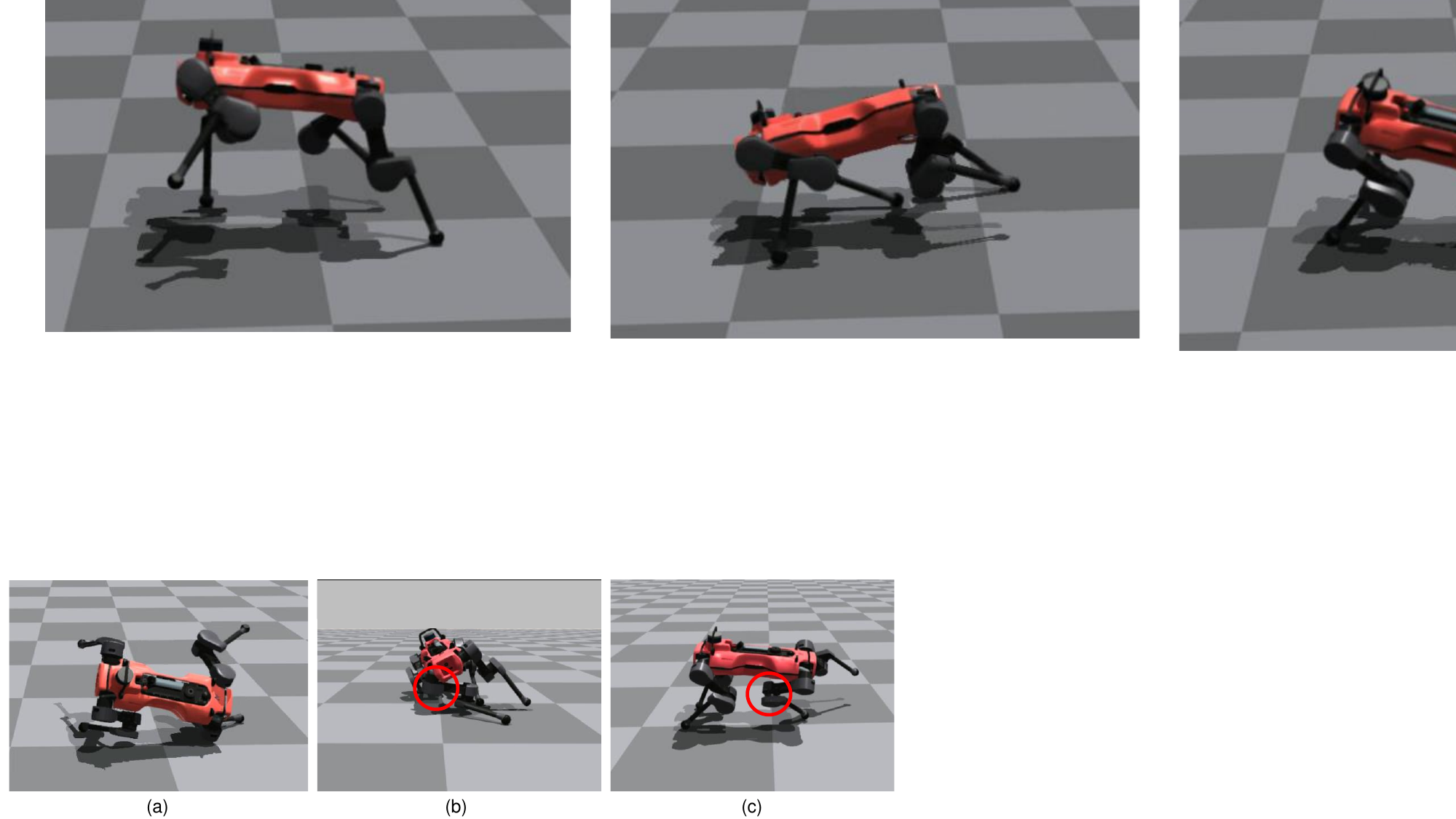}
    \caption{Results of learned diverse failure scenarios that could cause damage or prompt termination in the actual robot: (a) Falling over; (b) Self-collision with the base hitting a front leg; (c) Knee motor exceeding the torque limit due to the configuration and a single foot touchdown.}
    \label{figure:diverse-failure}
  \end{center}
  \vspace{-3mm}
\end{figure}

To be specific, we adjust the weights of different reward terms in Eq.~\ref{eq:reward}, as listed in Table~\ref{TAB:diverse-weight}. With different reward weights, we can learn different adversary behaviors, as shown in Fig.~\ref{figure:diverse-failure} where we attacked the command space and the observation space.

\begin{table}[t]
\centering
\caption{Diverse Adversarial Attacks and Corresponding Weights}
\label{TAB:diverse-weight}
\begin{tabular}{c|ccc}
\hline
\textbf{} & \textbf{Falling over} & \textbf{Self collision} & \textbf{Torque over limits} \\ \hline
$c_{orient}$ & 0.1                   & 0                       & 0                           \\ \hline
$c_{shake}$  & 0.05                  & 0                       & 0                           \\ \hline
$c_{torque}$ & 5.0                   & 0                       & 20.0                        \\ \hline
$\mathds{1}_{alive}$ & 1.0           & 1.0                     & 1.0                         \\ \hline
\end{tabular}
\end{table}

\begin{figure}[t]
  \begin{center}
    \includegraphics[clip,  bb= 0 0 600 540,  width=1.0\columnwidth]{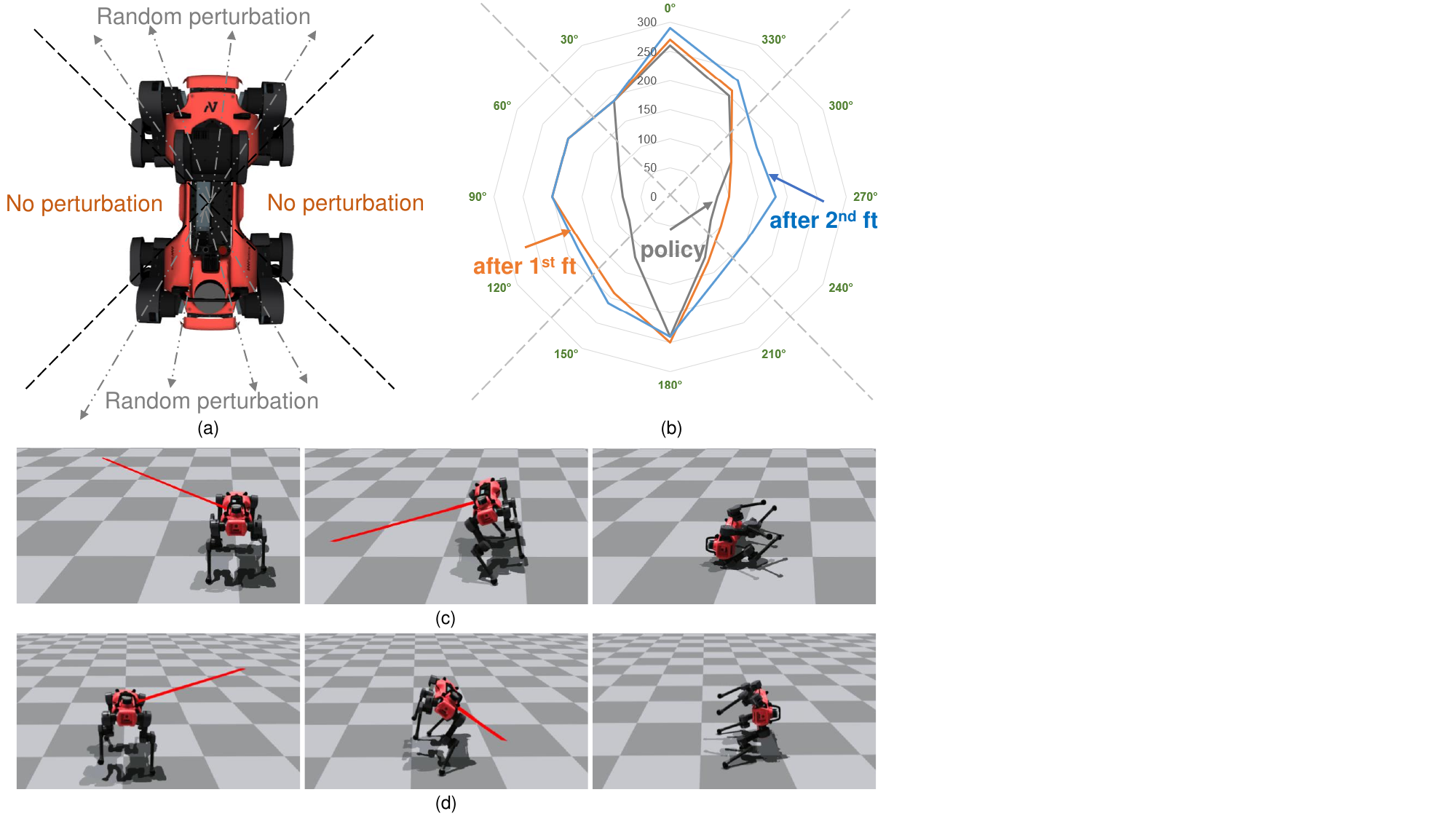}
    \caption{\Revise{Example to illustrate the effects of robustification through multiple rounds of attacking and finetuning (ft). (a) Initial setup: The domain randomization is deliberately biased to exclude lateral forces. (b) Minimal constant force perturbations required to induce failure: Comparisons among the initial policy (gray), the policy finetuned after the first adversarial attack (orange), and the policy further refined after both adversarial attacks (blue). (c) First attack findings: Reveals control policy weaknesses when the robot is pushed leftward. (d) Second attack outcomes: Exposes vulnerabilities when the robot is pushed rightward.
}}
    \label{figure:sequential-fine-tune-safeset}
  \end{center}
  \vspace{-3mm}
\end{figure}

\Revise{
\subsection{Illustrating effects of robustification: saving the unguarded}




We designed an experiment to illustrate the efficacy of our proposed methodology through multiple rounds of attacking and finetuning. To this end, we use the \textit{didactic policy} but deliberately bias the domain randomization during initial training: the external forces can only be sampled within $45^{\circ}$ from the $x$-axis in the robot’s sagittal plane (both forward and backward), as illustrated in Fig.~\ref{figure:sequential-fine-tune-safeset}(a).

As shown in Fig.~\ref{figure:sequential-fine-tune-safeset}(b), the initial policy is predictably vulnerable to lateral pushes according to our push tests. We trained the first adversary, which successfully identified and exploited weaknesses when pushing from the left, causing the robot to fall as Fig.~\ref{figure:sequential-fine-tune-safeset}(c). After finetuning with the first adversary, the controller became more robust against leftward pushes but remained vulnerable to rightward pushes. Subsequently, a second adversary was trained, which pinpointed and exploited weaknesses from the right, revealing further vulnerabilities in the controller as Fig.~\ref{figure:sequential-fine-tune-safeset}(d). By finetuning with the second adversary, the policy was significantly enhanced, showing improved resistance to pushes from both sides.

This example underscores the utility of adversarial attacks, particularly when the domain randomization employed fails to sufficiently cover the range of scenarios necessary for robust robot operation. 
}

\subsection{Bonus: attacking MPC}

\begin{figure}[t]
  \begin{center}
    \includegraphics[clip,  bb= 0 0 360 190,  width=0.8\columnwidth]{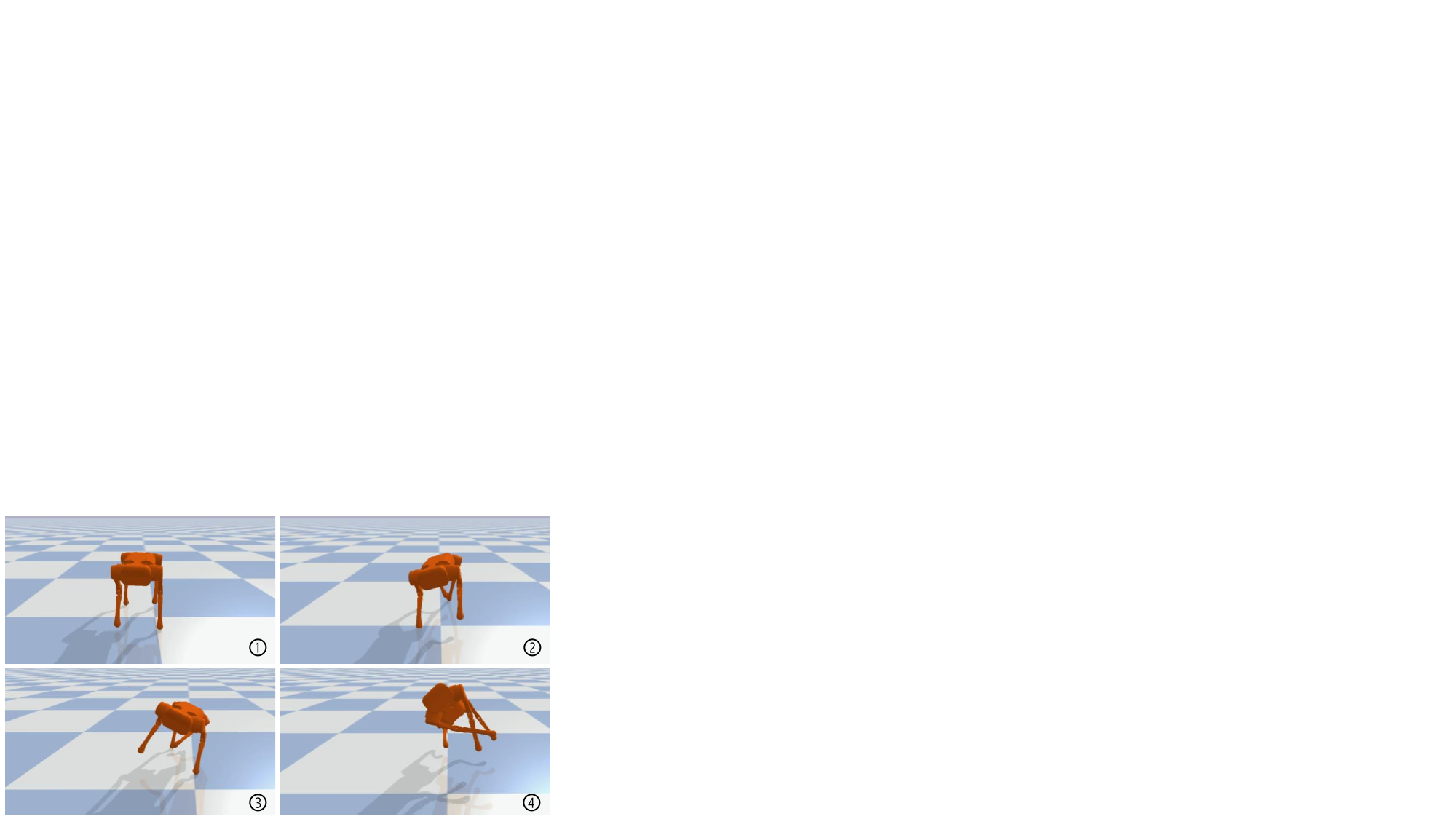}
    \caption{Results of learned adversarial attacks on a MPC controller in the simulation. Our attack is in command and observation space with the same range and reward setting used for the \textit{Miki policy} attacking. }
    \label{figure:mpc-attack}
  \end{center}
  \vspace{-3mm}
\end{figure}

Principally, our proposed method is not limited to the specific RL-based controller or any robot type. Here, we demonstrate that the model predictive controller (MPC) can also be subjected to learning-based attacks which are time-variant and state-dependent. To this end, we conducted successful attacks on an open-source MPC~\cite{kim2018mpc} for the Unitree A1 robot
\footnote{The controller was developed by \citet{kim2018mpc} for mini cheetah, and we use the adapted implementation for A1 by \citet{chen2023mpc_rl}.}, as shown in Fig.~\ref{figure:mpc-attack}

\section{Conclusion and limitations} 
\label{sec:conclusion}

In this paper, we present a learning-based adversarial attack framework targeting neural network based quadrupedal locomotion controllers. The learned adversaries can effectively uncover the vulnerabilities of the controllers, including the state-of-the-art DARPA-winning locomotion policy. 
By formulating the problem into a Markov Decision Process, a reinforcement learning method with Lipschitz regularization is employed to efficiently learn realistic sequential attacks.
The efficacy of our method is validated both in simulation and on the actual robot, with consistent behaviors between the simulation and the real world.

Furthermore, we demonstrate how learned adversarial attacks can be utilized to enhance the robustness of the controllers and potentially act as an analytical tool for robustness assessment. Besides, we provide extensive analyses and insights on robustness. Some key takeaways are: 
\begin{enumerate}
    \item Domain randomization is not sufficient to ensure the robustness against learned attacks;
    \item Multi-modal attacks are stronger than unimodal ones;
    \item Adversaries with exteroception can leverage terrains;
    \item Human intelligence on attacks does not diminish the significance of our learning-based approach.
\end{enumerate}
We also study how to diversify the learned attacks and whether our proposed method is applicable to other controllers such as model predictive controllers.

We hope this paper can raise awareness among legged robot researchers regarding the vulnerability in neural controllers and the critical need for comprehensive safety verification. By implementing the effective verification process, the safety and trustworthiness of robot controllers can be improved, paving the way for their secure large-scale deployment.

\subsection{Limitations}
Regarding the observation space, this paper attacks solely on the proprioception. 
However, for perceptive locomotion policies, the exteroceptive observations are also likely to be vulnerable, given their high dimensionality (e.g., $208$-dim for the \textit{Miki policy}). 
The proposed RL-based method might not be the most effective approach to output such high-dimensional sequences, and we will explore generative methods in the future.

Another limitation is the absence of theoretical guarantees to cover all of the weaknesses. Although we showcase how we can learn diverse attacks in Sec.~\ref{subsec:diverse}, investigating new techniques that maintain diversity from the machine learning community can be valuable.

Furthermore, our proposed method can serve as a tool to compare different frameworks and methods. However, ensuring fairness in these comparisons necessitates substantial effort to formally verify that the policies assessed are 'optimal' for their respective reinforcement learning algorithms and robotic configurations. Currently, the diversity of robotic platforms and sensor settings complicates direct comparisons. We plan to tackle these challenges through more rigorous comparisons in our future work.

\section{acknowledgment}

The authors would like to thank Prof. Kaiqing Zhang and Julian Nubert for the insightful discussion.

This work was supported by the European Research Council (ERC) under the European Union’s Horizon 2020 research and innovation program (grant agreement No. 866480, 780883), 
This project has received funding from the European Union’s Horizon 2020 research and innovation programme under grant agreement No. 101016970.

This research was partially supported by the ETH AI Center.


\bibliographystyle{plainnat}
\bibliography{references}

\end{document}